%% file: 3DBPP.tex
\def\year{2021}\relax
\pdfoutput=1
\documentclass[letterpaper]{article} 
\usepackage{aaai21}  
\usepackage{times}  
\usepackage{helvet} 
\usepackage{courier}  
\usepackage[hyphens]{url}  
\usepackage{graphicx} 
\usepackage{amssymb}
\usepackage{pifont}
\newcommand{\cmark}{\ding{51}}%
\newcommand{\xmark}{\ding{55}}%
\usepackage{multirow}
\usepackage{subfigure}
\usepackage{wrapfig}
\usepackage{natbib}  
\usepackage{caption} 
\input{ourcommands}

\usepackage{color}
\definecolor{green}{rgb}{0, 0.5, 0}
\definecolor{orange}{rgb}{0.8, 0.6, 0.2}
\definecolor{red}{rgb}{1.0, 0.0, 0.0}
\definecolor{teal}{rgb}{0.0, 0.4, 0.4}
\definecolor{purple}{rgb}{0.65,0,0.65}
\definecolor{saffron}{rgb}{0.95,0.75,0.2}
\definecolor{turquoise}{rgb}{0.0,0.5,0.5}

\newcommand{\cy}[1]{{\color{black}{#1}}}
\newcommand{\kx}[1]{{\color{black}#1}}

\newcommand{\nonl}{\renewcommand{\nl}{\let\nl\oldnl}}

\title{Online 3D Bin Packing with Constrained Deep Reinforcement Learning}
\author{Hang Zhao\textsuperscript{\rm 1}, \quad Qijin She\textsuperscript{\rm 1}, \quad Chenyang Zhu\textsuperscript{\rm 1}, \quad Yin Yang\textsuperscript{\rm 2}, \quad Kai Xu\textsuperscript{\rm 1}\thanks{Hang Zhao and Qijin She are co-first authors. Kai Xu is the corresponding author (kevin.kai.xu@gmail.com).}\\
\textsuperscript{\rm 1}National University of Defense Technology, \quad \textsuperscript{\rm 2}Clemson University\\}

\begin{document}

\maketitle

\input{abstract}
\input{intro}

\input{related}

\input{method}

\input{reorder}
\input{result}

\input{conclusion}

\section*{Acknowledgments}
We thank the anonymous PC, AC and extra reviewers for their insightful comments and valuable suggestions.
We are also grateful to the colleagues of SpeedBot Robotics for their help on real robot test.
Thanks also go to Chi~Trung~Ha for providing the source code of their work~\cite{ha2017online}.
This work was supported in part by the National Key Research and Development Program of China (No. 2018AAA0102200), NSFC (62002376, 62002375, 61532003, 61572507, 61622212) and NUDT Research Grants
(No. ZK19-30).

\bibliography{reference}

\newpage
\appendix
\input{supp}

\end{document}

%% file: ourcommands.tex
\usepackage{mathtools}
\usepackage{amsmath}

\usepackage{amssymb}
\usepackage{amsfonts}
\usepackage{amsopn}
\usepackage{graphicx} 
\usepackage{textcomp}
\usepackage{xfrac}
\usepackage{bbm}
\usepackage{overpic}
\usepackage{subfig}
\usepackage{wrapfig}
\usepackage[ruled,vlined,linesnumbered]{algorithm2e}
\usepackage{multirow}

\usepackage{color}
\definecolor{green}{rgb}{0, 0.5, 0}
\definecolor{orange}{rgb}{0.8, 0.6, 0.2}
\definecolor{red}{rgb}{1.0, 0.0, 0.0}
\definecolor{teal}{rgb}{0.0, 0.4, 0.4}
\definecolor{purple}{rgb}{0.65,0,0.65}
\definecolor{saffron}{rgb}{0.95,0.75,0.2}
\definecolor{turquoise}{rgb}{0.0,0.5,0.5}
\definecolor{brown}{rgb}{0.5, 0.16, 0.16}

\usepackage{overpic}
\usepackage{currfile} 

\newlength\savedwidth
\newcommand\whline[1]{\noalign{\global\savedwidth\arrayrulewidth
		\global\arrayrulewidth #1} %
	\hline
	\noalign{\global\arrayrulewidth\savedwidth}}

\newcommand{\supl}[1]{{\color{black}\emph{#1}}}

\definecolor{lightgray}{rgb}{0.6, 0.6, 0.6}

\usepackage[normalem]{ulem}

\newcommand{\hidecomment}[1]{}

\usepackage{xspace}



%% file: abstract.tex
\begin{abstract}
We solve a challenging yet practically useful variant of 3D Bin Packing Problem (3D-BPP). In our problem, the agent has limited information about the items to be packed into a \cy{single} bin, and an item must be packed immediately after its arrival without buffering or readjusting.
The item's placement also subjects to the constraints of order dependence and physical stability.
We formulate this \emph{online 3D-BPP} as a constrained Markov decision process (CMDP). To solve the problem, we propose an effective and easy-to-implement constrained deep reinforcement learning (DRL) method under the actor-critic framework. In particular, we introduce a \emph{prediction-and-projection} scheme: The agent first predicts a feasibility mask for the placement actions as an auxiliary task and then uses the mask to modulate the action probabilities output by the actor during training. Such supervision and projection facilitate the agent to learn feasible policies very efficiently. Our method can be easily extended to handle lookahead items, multi-bin packing, and item re-orienting. We have conducted extensive evaluation showing that the learned policy significantly outperforms the state-of-the-art methods. A preliminary user study even suggests that our method might attain a human-level performance.
\end{abstract} 

%% file: intro.tex
\section{Introduction} \label{sec:intro}

As a classic NP-hard problem, the bin packing problem (1D-BPP) seeks for an assignment of a collection of items with various weights to bins. The optimal assignment houses all the items with the fewest bins such that the total weight of items in a bin is below the bin's capacity $c$~\cite{korte2012bin}.
In its 3D version i.e., 3D-BPP~\cite{martello2000three}, an item $i$ has a 3D ``weight'' corresponding to its length, $l_i$, width $w_i$, and height $h_i$. Similarly, $c$ is also in 3D including $L \geq l_i$, $W \geq w_i$, and $H \geq h_i$. It is assumed that $l_i, w_i, h_i, L, W, H \in Z^+$ are positive integers. Given the set of items $\mathcal{I}$, we would like to pack all the items into as few bins as possible. Clearly, 1D-BPP is a special case of its three dimensional counter part -- as long as we constrain $h_i = H$ and $w_i = W$ for all $i\in\mathcal{I}$, a 3D-BPP instance can be relaxed to a 1D-BPP. Therefore, 3D-BPP is also highly NP-hard~\cite{man1996approximation}.

\begin{figure}\centering
	\includegraphics[width = 0.82\linewidth]{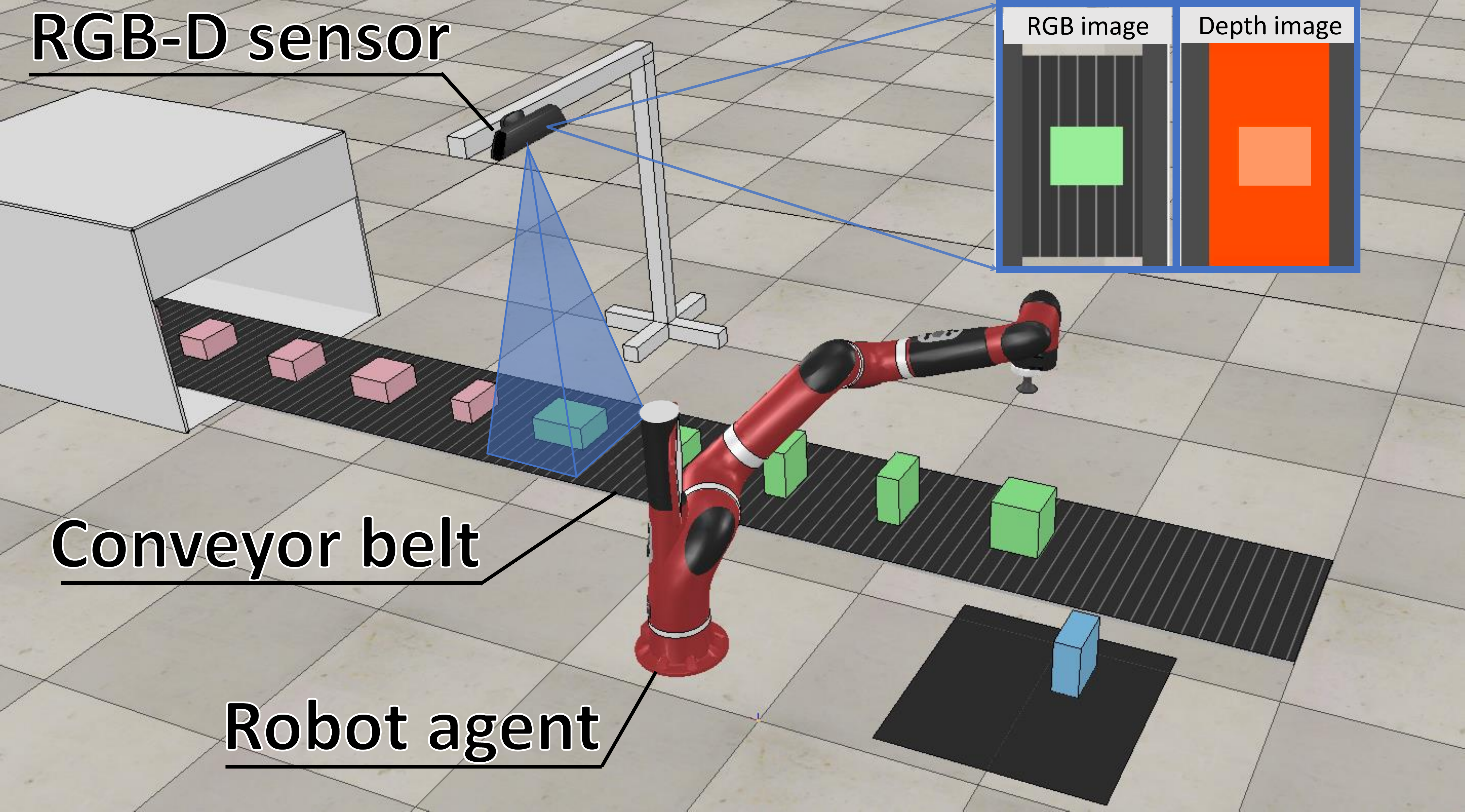}\vspace{-6pt}
	\caption{Online 3D-BPP, where the agent observes only a limited numbers of lookahead items (shaded in green), is widely useful in logistics, manufacture, warehousing etc.}\label{fig:teaser}\vspace{-16pt}
\end{figure}

Regardless of its difficulty, the bin packing problem turns out to be one of the most needed academic problems~\cite{skiena1997stony} (the second most needed, only after the suffix tree problem) as many real-world challenges could be much more efficiently handled if we have a good solution to it. A good example is large-scale parcel packaging in modern logistics systems (Figure.~\ref{fig:teaser}), where parcels are mostly in regular cuboid shapes, and we would like to collectively pack them into rectangular bins of the standard dimension. Maximizing the storage use of bins effectively reduces the cost of inventorying, wrapping, transportation, and warehousing. While being strongly NP-hard, 1D-BPP has been extensively studied. With the state-of-the-art computing hardware, big 1D-BPP instances (with about $1,000$ items) can be exactly solved within tens of minutes~\cite{delorme2016bin} using e.g., integer linear programming (ILP)~\cite{schrijver1998theory}, and good approximations can be obtained within milliseconds. On the other hand 3D-BPP, due to the extra complexity imposed, is relatively less explored. Solving a 3D-BPP of moderate size exactly (either using ILP or branch-and-bound) is much more involved, and we still have to resort to heuristic algorithms~\cite{crainic2008extreme,karabulut2004hybrid}.

Most existing 3D-BPP literature assumes that the information of all items is known while does not take physical stability into consideration, and the packing strategies allow backtracking i.e., one can always repack an item from the bin in order to improve the current solution~\cite{martello2000three}. In practice however, we do not know the information of all items. For instance see Figure~\ref{fig:teaser}, where a robot works beside a bin, and a conveyor forwards parcels sequentially. The robot may only have the vision of several upcoming items (similar to Tetris), and an item must be packed within a given time period after its arrival. It is costly and inefficient if the robot frequently unloads and readjusts parcels in packed bins. Such constraints further complicate 3D-BPP in its real-world applications.

As an echo to those challenges, we design a deep reinforcement learning algorithm for 3D-BPP. To maximize the applicability, we carefully accommodate restrictions raised in its actual usage.
For instance, we require item placement satisfying order dependence and not inducing instable stacking.
An item is immediately packed upon its arrival, and no adjustment will be permitted after it is packed. To this end, we opt to formulate our problem as a constrained Markov decision process (CMDP)~\cite{altman1999constrained} and propose a constrained DRL approach based on the on-policy actor-critic framework~\cite{mnih2016asynchronous,wu2017scalable}.
In particular, we introduce a \emph{prediction-and-projection} scheme for the training of constrained DRL.
The agent first predicts a feasibility mask for the placement actions as an auxiliary task.
It then uses the mask to modulate the action probabilities output by the actor.
These supervision and projection enable the agent to learn feasible policy very efficiently.
We also show that our method is general with the ability to handle lookahead items, multi-bin packing, and item re-orienting. With a thorough test and validation, we demonstrate that our algorithm outperforms existing methods by a noticeable margin. It even demonstrates a human-level performance in a preliminary user study.

%% file: related.tex
\section{Related Work}\label{sec:related}
\paragraph{1D-BPP} is one of the most famous problems in combinatorial optimization, and related literature dates back to the sixties~\cite{kantorovich1960mathematical}. Many variants and generalizations of 1D-BPP arise in practical contexts such as the cutting stock problem (CSP), in which we want to cut bins to produce desired items of different weights, and minimize the total number of bins used. A comprehensive list of bibliography on 1D-BPP and CSP can be found in~\cite{sweeney1992cutting}. Knowing to be strongly NP-hard, most existing literature focuses on designing good heuristic and approximation algorithms and their worst-case performance analysis~\cite{coffman1984approximation}. For example, the well-known greedy algorithm, the next fit algorithm (NF) has a linear time complexity of $O(N)$ and its \emph{worst-case performance ratio} is 2 i.e. NF needs at most twice as many bins as the optimal solution does~\cite{de1981bin}. The first fit algorithm (FF) allows an item to be packed into previous bins that are not yet full, and its time complexity increases to $O(N \log N)$. The best fit algorithm (BF) aims to reduce the residual capacity of all the non-full bins. Both FF and BF have a better worst-case performance ratio of $\frac{17}{10}$ than NF~\cite{johnson1974worst}. Pre-sorting all the items yields the off-line version of those greedy strategies sometimes also known as the \emph{decreasing} version~\cite{martello1990knapsack}. While straightforward, NF, FF, and BF form a foundation of more sophisticated approximations to 1D-BPP (e.g. see~\cite{karmarkar1982efficient}) or its exact solutions~\cite{martello1990lower,scholl1997bison,labbe1995exact,delorme2016bin}. We also refer the reader to \texttt{BPPLib} library~\cite{delorme2018bpplib}, which includes the implementation of most known algorithms for the 1D-BPP problem.

\vspace{-8pt}
\paragraph{2D- and 3D-BPP} are natural generalization of the original BPP. Here, an item does not only have a scalar-valued weight but a high-dimension size of width, height, and/or depth. The main difference between 1D- and 2D-/3D- packing problems is the verification of the feasibility of the packing, i.e. determining whether an accommodation of the items inside the bin exists such that items do not interpenetrate and the packing is within the bin size. The complexity and the difficulty significantly increase for high-dimension BPP instances. In theory, it is possible to generalize exact 1D solutions like MTP~\cite{martello1990lower} or branch-and-bound~\cite{delorme2016bin} algorithms to 2D-BPP~\cite{martello1998exact} and 3D-BPP~\cite{martello2000three}. However according to the timing statistic reported in~\cite{martello2000three}, exactly solving 3D-BPP of a size matching an actual parcel packing pipeline, which could deal with tens of thousand parcels, remains infeasible. Resorting to approximation algorithms is a more practical choice for us. Hifi et al.~\shortcite{hifi2010linear} proposed a mixed linear programming algorithm for 3D-BPP by relaxing the integer constraints in the problem. Crainic et al.~\shortcite{crainic2008extreme} refined the idea of corner points~\cite{martello2000three}, where an upcoming item is placed to the so-called \emph{extreme points} to better explore the un-occupied space in a bin.
\kx{Heuristic local search iteratively improves an existing packing by searching within a neighbourhood function over the set of solutions. There have been several strategies in designing fast approximate algorithms, e.g., guided local search~\cite{faroe2003guided}, greedy search~\cite{de2003greedy}, and tabu search~\cite{lodi1999approximation,crainic2009ts2pack}. Similar strategy has also been adapted to Online BPP~\cite{ha2017online,wang2016benchmarking}. In contrast, genetic algorithms leads to better solutions as a global, randomized search~\cite{li2014genetic,takahara2005evolutionary}.}

\vspace{-8pt}
\paragraph{Deep reinforcement learning (DRL)}
has demonstrated tremendous success in learning complex behaviour skills and solving challenging control tasks with high-dimensional raw sensory state-space~\cite{lillicrap2015continuous,mnih2015,mnih2016asynchronous}.
\kx{The existing research can largely be divided into two lines: on-policy methods~\cite{schulman2017proximal, wu2017scalable, zhao2018triangle} and off-policy ones~\cite{mnih2015, wang2015dueling, barth2018distributed}.
On-policy algorithms optimize the policy with agent-environment interaction data sampled from the current policy.
While lacking the ability of reusing old data makes them less data efficient, updates calculated by on-policy data lead to stable optimization.
In contrast, off-policy methods are more data-efficient but less stable.
In our problem, agent-environment interaction data is easy to obtain (in $2000$FPS), thus data efficiency is not our main concern. We base our method on the on-policy actor-critic framework.
In addition, we formulate online 3D-BPP as constrained DRL and solve it by projecting the trajectories sampled from the actor to the constrained state-action space, instead of resorting to more involved constrained policy optimization~\cite{achiam2017constrained}.}

\begin{figure*}
  \centering
  \includegraphics[width=\linewidth]{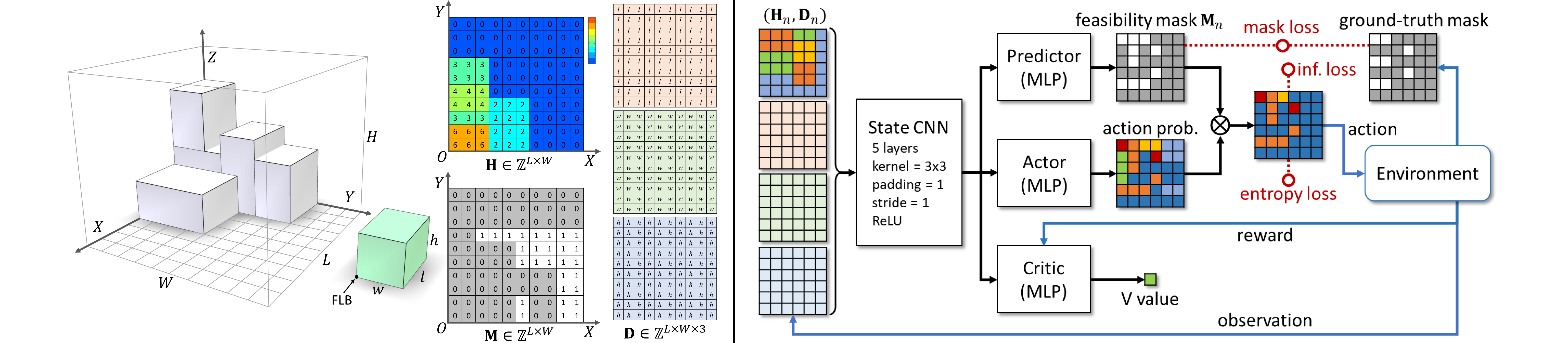}
  \caption{Left: The environment state of the agent includes the configuration of the bin (the grey boxes) and the size of the next item to be packed (green box). The bin configuration is parameterized as a height map $\mathbf{H}$ over a $L \times W$ grid. The feasibility mask $\mathbf{M}$ is a binary matrix of size $L \times W$ indicating the placement feasibility at each grid cell. The three dimensions of the next item are stored into a $L \times W \times 3$ tensor $\mathbf{D}$.
  Right: The network architecture (the three losses other than the standard actor and critic losses are shown in red color).}
 \label{fig:3D-BPP}\vspace{-13pt}
\end{figure*}

\vspace{-8pt}
\paragraph{RL for combinatorial optimization}
\kx{
has a distinguished history~\cite{gambardella1995ant,zhang2000solving} and is still an active direction with especially intensive focus on TSP~\cite{bello2016neural}.
Early attempts strive for heuristics selection using RL~\cite{nareyek2003choosing}.
Bello et al.~\shortcite{bello2016neural} combined \emph{RL pretraining} and \emph{active search} and demonstrated that RL-based optimization outperforms supervised learning framework when tackling NP-hard combinatorial problems.
Recently, Hu et al.~\shortcite{hu2017solving} proposed a DRL solution to 3D-BPP. Laterre et al.~\shortcite{laterre2018ranked} introduced a rewarding strategy based on self-play. Different from ours, these works deal with an offline setting where the main goal is to find an optimal sequence of items inspired by the Pointer Network~\cite{vinyals2015pointer}.
}

%% file: method.tex
\section{Method}\label{sec:method}
In online 3D-BPP, the agent is agnostic on $l_i$, $w_i$ or $h_i$ of all the items in $\mathcal{I}$ -- only immediately incoming ones $\mathcal{I}_o\subset\mathcal{I}$ are observable. As soon as an item arrives, we pack it into the bin, and no further adjustment will be applied. 
As the complexity of BPP decreases drastically for bigger items, we further constrain the sizes of all items to be $l_i \leq L/2$, $w_i \leq W/2$, and $h_i \leq H/2$. We start with our problem statement under the context of DRL and the formulation based on constrained DRL. We show how we solve the problem via predicting action feasibility in the actor-critic framework.

%
\subsection{Problem statement and formulation}
The 3D-BPP can be formulated as a Markov decision process, which is a tuple of $(\mathcal{S}, \mathcal{A}, P, R)$. $\mathcal{S}$ is the set of environment states; $\mathcal{A}$ is the action set; $R : \mathcal{S} \times \mathcal{A} \rightarrow \mathbb{R}$ is the reward function;
$P : \mathcal{S} \times \mathcal{A} \times \mathcal{S} \rightarrow [0,1]$ is the transition probability function. $P(s'|s,a)$ gives the probability of transiting from $s$ to $s'$ for given action $a$. Our method is model-free since we do not learn $P(s'|s,a)$. The policy $\pi : \mathcal{S} \rightarrow \mathcal{A}$ is a map from states to probability distributions over actions, with $\pi(a|s)$ denoting the probability of selecting action $a$ under state $s$. For DRL, we seek for a policy $\pi$ to maximize the accumulated discounted reward, $J(\pi)=E_{\tau \sim \pi}[\sum_{t=0}^{\infty}\gamma^tR(s_t,a_t)]$. Here, $\gamma \in [0,1]$ is the discount factor, and $\tau = (s_0,a_0,s_1,\ldots)$ is a trajectory sampled based on the policy $\pi$.


The environment state of 3D-BPP is comprised of two parts: the current configuration of the bin and the coming items to be placed. For the first part, we parameterize the bin through discretizing its bottom area as a $L \times W$ regular grid along length ($X$) and width ($Y$) directions, respectively.
We record at each grid cell the current height of stacked items, leading to a \emph{height map} $\mathbf{H}_n$ (see Figure~\ref{fig:3D-BPP}). Here, the subscript $n$ implies $n$ is the next item to be packed. Since all the dimensions are integers, $\mathbf{H}_n\in\mathbb{Z}^{L \times W}$ can be expressed as a 2D integer array. The dimensionality of item $n$ is given as  $\mathbf{d}_n = [l_n, w_n, h_n]^\top\in\mathbb{Z}^3$.
\kx{Working with integer dimensions helps to reduce the state/action space and accelerate the policy learning significantly. 
A spatial resolution of up to $30 \times 30$ is sufficient in many real scenarios.
}
Putting together, the current environment state can be written as $s_n = \{\mathbf{H}_n, \mathbf{d}_n, \mathbf{d}_{n+1},...,\mathbf{d}_{n+k-1}\}$. We first consider the case where $k=|\mathcal{I}_o|=1$, and name this special instance as BPP-$1$. In other words, BPP-$1$ only considers the immediately coming item $n$ i.e., $\mathcal{I}_o=\{ n \}$. We then generalize it to BPP-$k$ with $k>1$ afterwards.

\vspace{-10pt}
\paragraph{BPP-$1$}
In BPP-$1$, the agent places $n$'s front-left-bottom (FLB) corner (Figure~\ref{fig:3D-BPP} (left)) at a certain grid point or the loading position (LP) in the bin. For instance, if the agent chooses to put $n$ at the LP of $(x_n, y_n)$. This action is represented as $a_n = x_n + L \cdot y_n \in \mathcal{A}$, where the action set $\mathcal{A} = \{0, 1, \ldots, L \cdot W - 1\}$.
After $a_n$ is executed, $\mathbf{H}_n$ is updated by adding $h_n$ to the maximum height over all the cells covered by $n$:
$\mathbf{H}'_{n}(x, y) = h_{\text{max}}(x, y) + h_n$ for $x\in[x_n, x_n + l_n], y\in[y_n, y_n + w_n]$,
with $h_{\text{max}}(x, y)$ being the maximum height among those cells.
The state transition is deterministic: $P(\mathbf{H}|\mathbf{H}_n, a_n) = 1$ for $\mathbf{H} = \mathbf{H}'_{n}$ and $P(\mathbf{H}|\mathbf{H}_n, a_n) = 0$ otherwise.



During packing, the agent needs to secure enough space in the bin to host item $n$. Meanwhile, it is equally important to have $n$ statically equilibrated by the underneath at the LP so that all the stacking items are physically stable. Evaluating the physical stability at a LP is involved, taking into account of $n$'s center of mass, moment of inertia, and rotational stability~\cite{goldstein2002classical}. All of them are normally unknown as the mass distribution differs among items. To this end, we employ a conservative and simplified criterion. Specifically, a LP is considered \emph{feasible} if it not only provides sufficient room for $n$ but also satisfies any of following conditions with $n$ placed: 1) over $60\%$ of $n$'s bottom area and all of its four bottom corners are supported by existing items; or 2) over $80\%$ of $n$'s bottom area and three out of four bottom corners are supported; or 3) over $95\%$ of $n$'s bottom area is supported. We store the feasibility of all the LPs for item $n$ with a \emph{feasibility mask} $\mathbf{M}_n$, an $L \times W$ binary matrix (also see Figure~\ref{fig:3D-BPP}).


Since not all actions are allowed, our problem becomes a constrained Markov decision processes (CMDP)~\cite{altman1999constrained}.
Typically, one augments the MDP with an auxiliary cost function $C : \mathcal{S} \times \mathcal{A} \rightarrow \mathbb{R}$ mapping state-action tuples to costs, and require that the expectation of the accumulated cost should be bounded by $c_m$:  $J_C(\pi) = E_{\tau \sim \pi}[\sum_{t=0}^{\infty}\gamma_C^tC(s_t,a_t)] \leq c_m$.
%
%
%
Several methods have been proposed to solve CMDP based on e.g.,
algorithmic heuristics~\cite{uchibe2007constrained}, primal-dual methods~\cite{chow2017risk}, or constrained policy optimization~\cite{achiam2017constrained}.
While these methods are proven effective, it is unclear how they could fit for 3D-BPP instances, where the constraint is rendered as a discrete mask. In this work, we propose to exploit the mask $\mathbf{M}$ to guide the DRL training to enforce the feasibility constraint without introducing excessive training complexity.




\subsection{Network architecture}\label{sec:network_architecture}
We adopt the actor-critic framework with Kronecker-Factored Trust Region (ACKTR)~\cite{wu2017scalable}. It iteratively updates an actor and a critic module jointly. In each iteration, the actor learns a policy network that outputs the probability of each action (i.e., placing $n$ at the each LP). The critic trains a state-value network producing the value function.
We find through experiments that on-policy methods (such as ACKTR) lead to better performance than off-policy ones like SAC~\cite{haarnoja2018soft}; \supl{see a comparison in the supplemental material.}

\vspace{-10pt}
\paragraph{State input}
In the original ACKTR framework, both actor and critic networks take the raw state directly as input. In our implementation however, we devise a CNN, named \emph{state CNN}, to encode the raw state vector into features. To facilitate this, we ``stretch'' $\mathbf{d}_n$ into a \kx{three-channel tensor $\mathbf{D}_n \in \mathbb{Z}^{L \times W \times 3}$ so that each channel of $\mathbf{d}_n$ spans a $L \times W$ matrix with all of its elements being $l_n$, $w_n$ or $h_n$, respectively (also see Figure~\ref{fig:3D-BPP}).} Consequently, state $s_n = (\mathbf{H}_n, \mathbf{D}_n)$ becomes a $L \times W \times 4$ array (Figure~\ref{fig:3D-BPP} (right)).

\vspace{-10pt}
\paragraph{Reward}
We define a simplistic step-wise reward as the volumetric occupancy introduced by the current item: $r_n = 10 \times l_n \cdot w_n \cdot h_n / (L \cdot W \cdot H)$ for item $n$. When the current item is not placeable, its reward is zero and the episode ends. While the feasibility mask saves the efforts of exploring invalid actions, this step-wise reward directs the agent to place as many items as possible.
We find through comparison that this step-wise reward is superior than a termination one (e.g. the final space utilization); \supl{see supplemental material}.



\vspace{-10pt}
\paragraph{Feasibility constraints}
We devise a prediction-and-projection mechanism to enforce feasibility constraints.
First, we introduce an independent multilayer perceptron module, namely the \emph{mask predictor} (Figure~\ref{fig:3D-BPP} (right)), to predict the feasibility mask $\mathbf{M}_n$ for the item $n$. The predictor takes the state CNN features of the current state as the input and is trained with the ground-truth mask as the supervision. Next, we use the predicted mask to modulate the output, i.e., the probability
\begin{wrapfigure}{r}{0.48\linewidth}\centering
    \vspace{-12 pt}
	\hspace{-25 pt}\includegraphics[width=\linewidth]{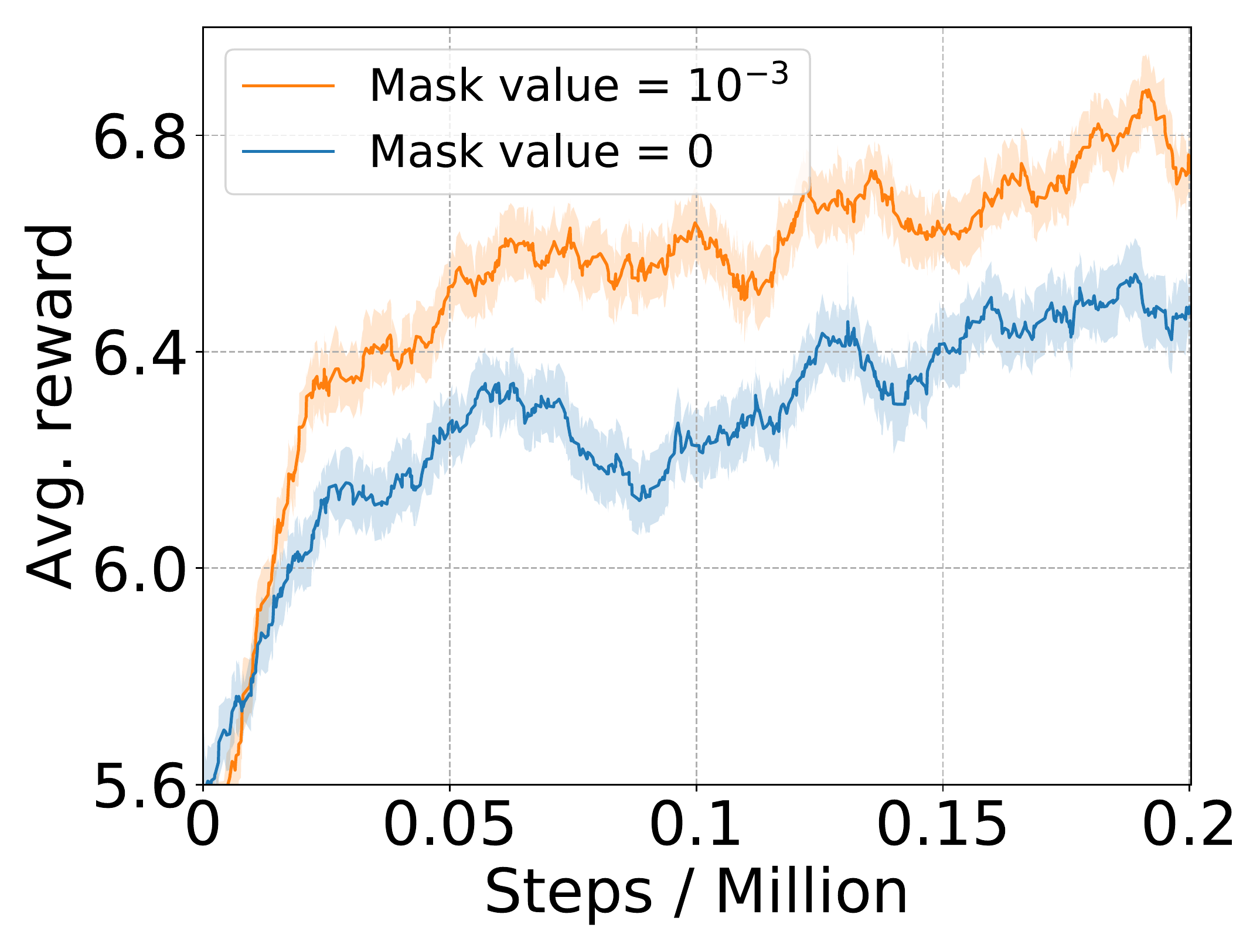}
	\label{fig:mask_comparison}
	\vspace{-14 pt}
\end{wrapfigure}
distribution of the actions.
In theory, if the LP at $(x, y)$ is infeasible for $n$, the corresponding probability $P(a_n= x + L\cdot y|s_n)$ should be set to $0$. However, we find that setting $P$ to a small positive quantity \cy{like $\epsilon = 10^{-3}$} works better in practice -- it provides a strong penalty to an invalid action but a smoother transformation beneficial to the network training. The inset shows that softening the mask-based modulation improves the training convergence.
To further discourage infeasible actions, we explicitly minimize the summed probability at all infeasible LPs: $E_{inf}=\sum P(a_n= x + L \cdot y|s_n)$, $\forall (x, y) | \mathbf{M}_n(x,y)=\epsilon$, which is plugged into the final loss function for training.
%

\vspace{-8pt}
\paragraph{Loss function}
Our loss function is defined as:
\begin{equation}\label{eq:loss_function}
L = \alpha \cdot L_{actor} + \beta \cdot L_{critic} + \lambda \cdot L_{mask} + \omega \cdot E_{inf} - \psi \cdot E_{entropy}.
\end{equation}
Here, $L_{actor}$ and $L_{critic}$ are the loss functions used for training the actor and the critic, respectively.
$L_{mask}$ is the MSE loss for mask prediction.
To push the agent to explore more LPs, we also utilize an action entropy loss
$E_{entroy} = \sum_{\mathbf{M}_n(x,y)=1}-P(a_n | s_n) \cdot \log \big(P(a_n | s_n)\big)$.
\setlength{\columnsep}{10 pt}
Note that the entropy is computed only over the set of all feasible actions whose LP satisfies $\mathbf{M}_n(x,y)=1$.
In this way, we stipulate the agent to explore only feasible actions.
We find through experiments that the following weights lead to consistently good performance
throughout our tests:
$\alpha=1$, $\beta=\lambda=0.5$, and $\omega=\psi=0.01$.



%% file: reorder.tex
\subsection{BPP-$k$ with $k=|\mathcal{I}_o|>1$}
\label{sec:reorder}
In a more general case, the agent receives the information of $k>1$ \emph{lookahead} items (i.e., from $n$ to $n+k-1$).
Obviously, the additional items inject more information to the environment state, which should be exploited in learning the policy $\pi(a_n|\mathbf{H}_n,\mathbf{d}_n,...,\mathbf{d}_{n+k-1})$.
One possible solution is to employ sequential modeling of the state sequence $(\mathbf{d}_n,...,\mathbf{d}_{n+k-1})$ using, e.g., recurrent neural networks. We found that, however, such state encoding cannot well inform the agent about the lookahead items during DRL training and yields limited improvement. Alternatively, we propose a search-based solution leveraging the height map $\mathbf{H}$ update and feasibility mask prediction.

The core idea is to condition the placement of the current item $n$ on the next $k-1$ ones.
Note that the actual placement of the $k$ items still follows the order of arrival. To make the current placement account for the future ones, we opt to ``hallucinate'' the placement of future items through updating the height map accordingly. Conditioned on the virtually placed future items, the decision for the current item could be globally more optimal. However, such virtual placement must satisfy the \emph{order dependence constraint} which stipulates that the earlier items should never be packed on top of the later ones.
In particular, given two items $p$ and $q$, $p < q$ in $\mathcal{I}_o$, if $q$ is (virtually) placed before $p$,
we require that the placement of $p$ should be spatially independent to the placement of $q$.
It means $p$ can never be packed at any LPs that overlap with $q$. This constraint is enforced by setting the height values in $\mathbf{H}$ at the corresponding LPs to $H$, the maximum height value allowed:  $\mathbf{H}_p(x,y) \leftarrow H$, for all $x\in[x_q, x_q + l_q]$ and $y\in[y_q, y_q + w_q]$.
Combining explicit height map updating with feasibility mask prediction, the agent utilizes the trained policy with the order dependence constraint satisfied implicitly.

\vspace{-8pt}
\paragraph{Monte Carlo permutation tree search}

\begin{figure}
	\includegraphics[width = \linewidth]{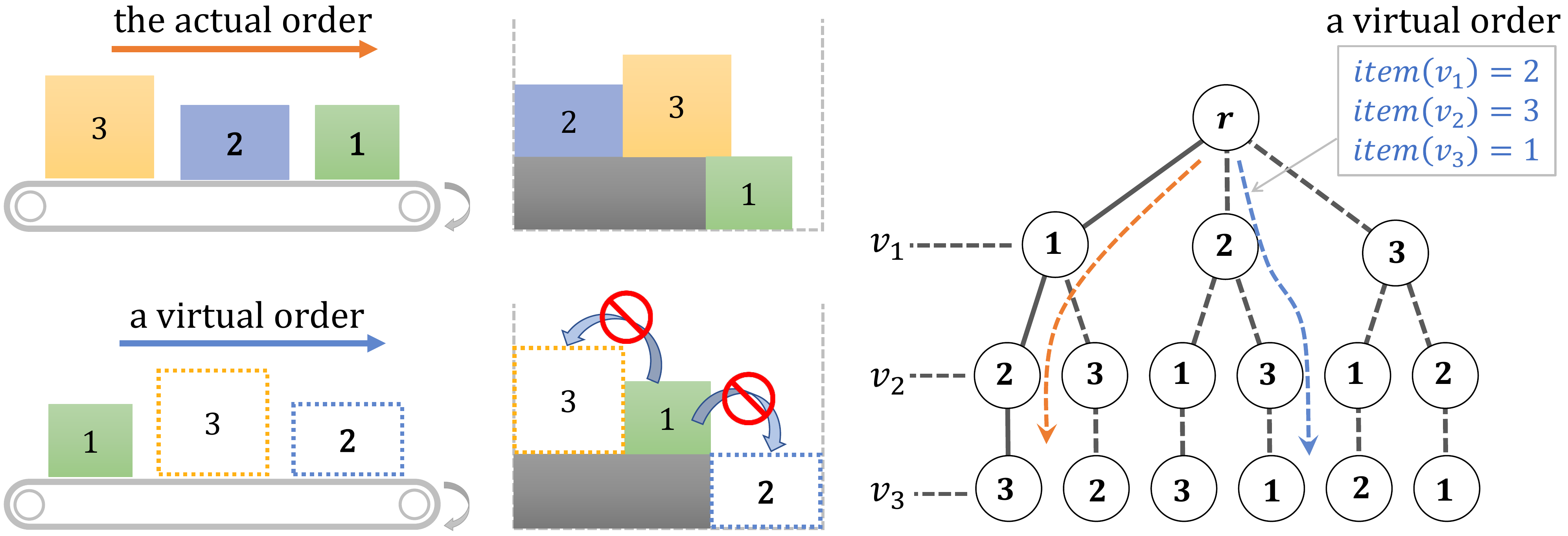}
	\caption{\kx{The permutation tree for $\mathcal{I}_o=\{1,2,3\}$. To find the best packing for item $1$, our method explores different virtual placing orders satisfying the order dependence constraint, e.g., $1$ cannot be placed on top of virtually placed $2$ or $3$.}}
	\label{fig:permutation_tree}\vspace{-15pt}
\end{figure}

We opt to search for a better $a_n$ through exploring the permutations of the sequence $(\mathbf{d}_n,...,\mathbf{d}_{n+k-1})$.
This amounts to a permutation tree search during which only the actor network test is conducted -- no training is needed.
\kx{
Figure~\ref{fig:permutation_tree} shows a $k$-level permutation tree: A path $(r, v_1, v_2,..., v_k)$ from the root to a leaf forms a possible permutation of the placement of the $k$ items in $\mathcal{I}_o$, where
$r$ is the (empty) root node and let $item(v_i)$ represent the $i$-th item being placed in the permutation.
Given two items $item(v_i) < item(v_j)$ meaning $item(v_i)$ arrives before $item(v_j)$ in the actual order.
If $i > j$ along a permutation path, meaning that $item(v_j)$ is virtually placed before $item(v_i)$,
we block the LPs corresponding to $item(v_j)$'s occupancy to avoid placing $item(v_i)$ on top of $item(v_j)$.

Clearly, enumerating all the permutations for $k$ items quickly becomes prohibitive with an $O(k!)$ complexity.
To make the search scalable, we adapt the Monte Carlo tree search (MCTS)~\cite{silver2017mastering} to our problem.
With MCTS, the permutation tree is expanded in a priority-based fashion through evaluating how promising a node would lead to the optimal solution. The latter is evaluated by sampling a fixed number of paths starting from that node and computing for each path a value summing up the accumulated reward and the critic value (``reward to go'') at the leaf ($k$-th level) node.
After search, we choose the action $a_n$ corresponding to the permutation with the highest path value.
\supl{Please refer to the supplemental material for more details on our adaptions of the standard MCTS.} MCTS allows a scalable lookahead for BPP-$k$ with a complexity of $O(km)$ where $m$ is the number of paths sampled.
}

%% file: result.tex
\section{Experiments}\label{sec:experiment}
We implement our framework on a desktop computer (\texttt{ubuntu 16.04}), which equips with an \texttt{Intel} \texttt{Xeon} \texttt{Gold} \texttt{5115} CPU @ 2.40 GHz, 64G memory, and a \texttt{Nvidia} \texttt{Titan} \texttt{V} GPU with 12G memory. The DRL and all other networks are implemented with \texttt{PyTorch}~\cite{paszke2019pytorch}. The model training takes about $16$ hours on a spatial resolution of $10\times 10$. The test time of BPP-$1$ model (no lookahead) is less than $10$~ms. \supl{Please refer to the supplemental material for more implementation details.}

\begin{table}
	\centering
	\scalebox{0.9}{
			\begin{tabular}{ccc|c|c}
				\whline{1.15pt}
				{MP} & {MC} & {FE} & {Space uti.} & {\# items} \\
				\whline{0.65pt}
				\xmark & \xmark & \xmark &  $7.82\%$ & $2.0$  \\
				\cmark & \xmark & \cmark & $27.9\%$ & $7.5$   \\
				\cmark & \cmark & \xmark & $63.7\%$ & $16.9$  \\
				 \xmark & \cmark & \cmark  &  $63.0\%$& $16.7$\\
				\cmark & \cmark & \cmark & $\bf{66.9\%}$ & $\bf{17.5}$          \\
				\whline{1.15pt}
			\end{tabular}}\vspace{-6pt}
\caption{This ablation study compares the space utilization and the total number of packed items with different combinations of MP, MC and FE, on the CUT-2 dataset.}\label{tab:ablation}\vspace{-12pt}
\end{table}

\vspace{-8pt}
\paragraph{Training and test set}
We set $L=W=H=10$ in our experiments with 64 pre-defined item dimensions ($|\mathcal{I}|=64$).
\supl{Results with higher spatial resolution are given in the supplemental material.}
We also set $l_i \leq L/2$, $w_i \leq W/2$ and $h_i \leq H/2$ to avoid over-simplified scenarios. The training and test sequence is synthesized by generating items out of $\mathcal{I}$, and the total volume of items should be equal to or bigger than bin's volume. We first create a benchmark called \textbf{RS} where the sequences are generated by sampling items out of $\mathcal{I}$ randomly.
A disadvantage of the random sampling is that the optimality of a sequence is unknown (unless performing a brute-force search). Without knowing whether the sequence would lead to a successful packing, it is difficult to gauge the packing performance with this benchmark.


\setlength{\columnsep}{10 pt}
\begin{wrapfigure}{r}{0.24\linewidth}
	\includegraphics[width = \linewidth]{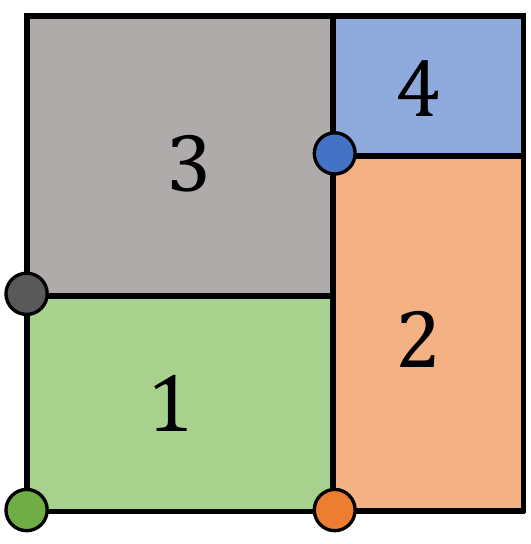}
    \vspace{-15 pt}
\vspace{-10 pt}
\end{wrapfigure}
Therefore, we also generate training sequences via \emph{cutting stock}~\cite{gilmore1961linear}. Specifically, items in a sequence are created by sequentially ``cutting'' the bin into items of the pre-defined 64 types so that we understand the sequence may be perfectly packed and restored back to the bin. There are two variations of this strategy. \textbf{CUT-1}: After the cutting, we sort resulting items into the sequence based on $Z$ coordinates of their FLBs, from bottom to top. If FLBs of two items have the same $Z$ coordinate, their order in the sequence is randomly determined. \textbf{CUT-2}: The cut items are sorted based on their stacking dependency: an item can be added to the sequence only after all of its supporting items are there.
A 2D toy example is given in the inset figure with FLB of each item highlighted.
Under CUT-1, both $\{1,2,3,4\}$ and $\{2,1,3,4\}$ are valid item sequences. If we use CUT-2 on the other hand,
$\{1,3,2,4\}$ and $\{2,4,1,3\}$ would also be valid sequences as the placement of $3$ or $4$ depends on $1$ or $2$.
For the testing purpose, we generate 2,000 sequences using RS, CUT-1, and CUT-2 respectively. The performance of the packing algorithm is quantitated with space utilization (\emph{space uti.}) and the total number of items packed in the bin (\emph{\# items}).

\supl{In the supplemental material, we provide visual packing results on all three datasets,
as well as an evaluation of model generalization across different datasets.}
\supl{Animated results can be found in the accompanying video.}

\begin{figure}
	\centering
	\includegraphics[width = \linewidth]{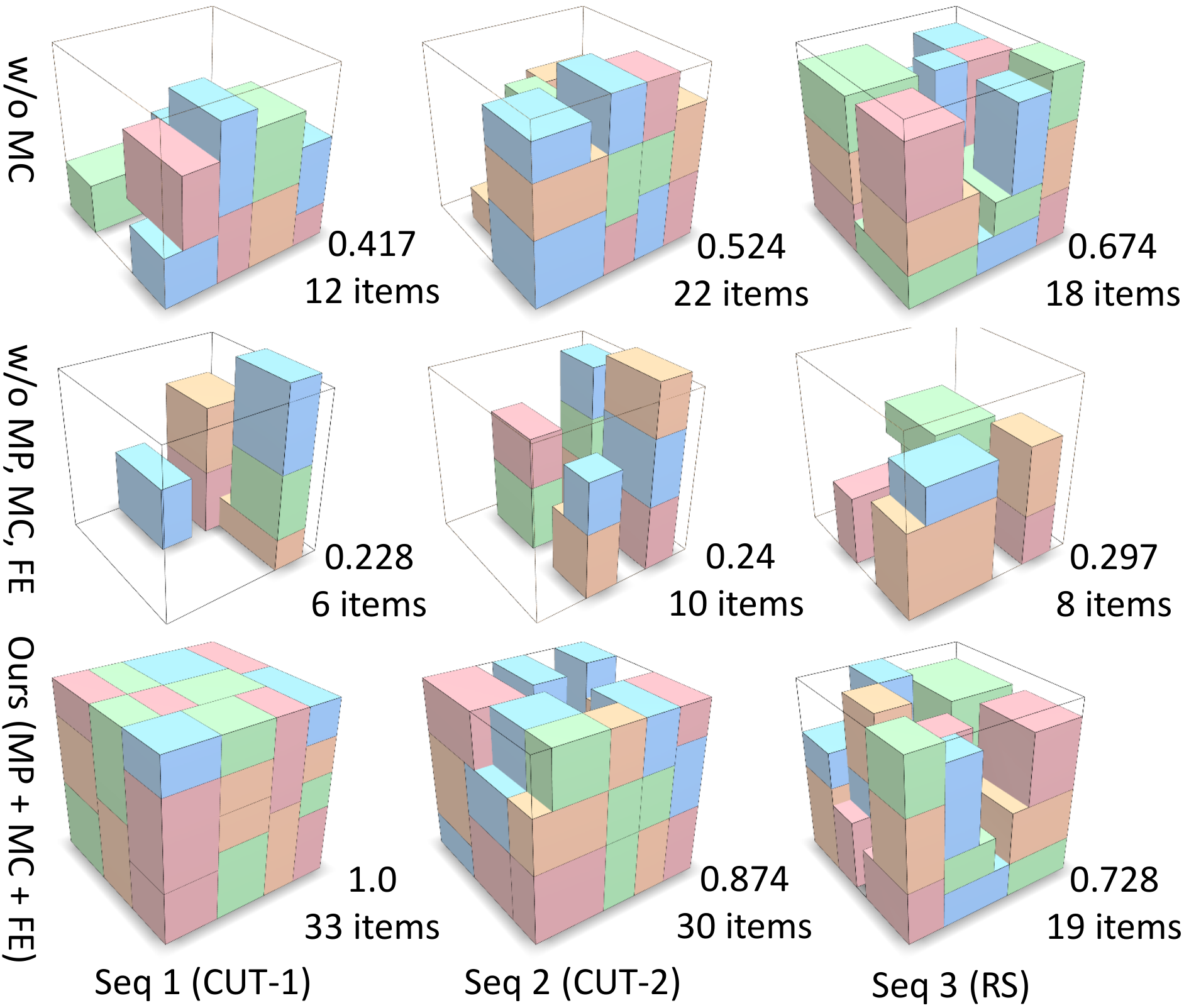} 
	\caption{Packing results in the ablation study. The numbers beside each bin are \emph{space uti.} and \emph{\# items}.}\label{fig:visual_ablation}\vspace{-16pt}
\end{figure}

%
%
%
%


\vspace{-10 pt}
\paragraph{Ablation study}
\kx{
Table~\ref{tab:ablation} reports an ablation study. From the results,
we found that the packing performance drops significantly if we do not incorporate the feasibility mask prediction (MP) during the training. The performance is impaired if the mask constraint (MC) is not enforced with our projection scheme. The feasibility-based entropy (FE) is also beneficial for both the training and final performance.
Figure~\ref{fig:visual_ablation} demonstrates the packing results visually for different method settings.
}

\begin{figure}[t!]\centering
    \begin{overpic}[width = \linewidth]{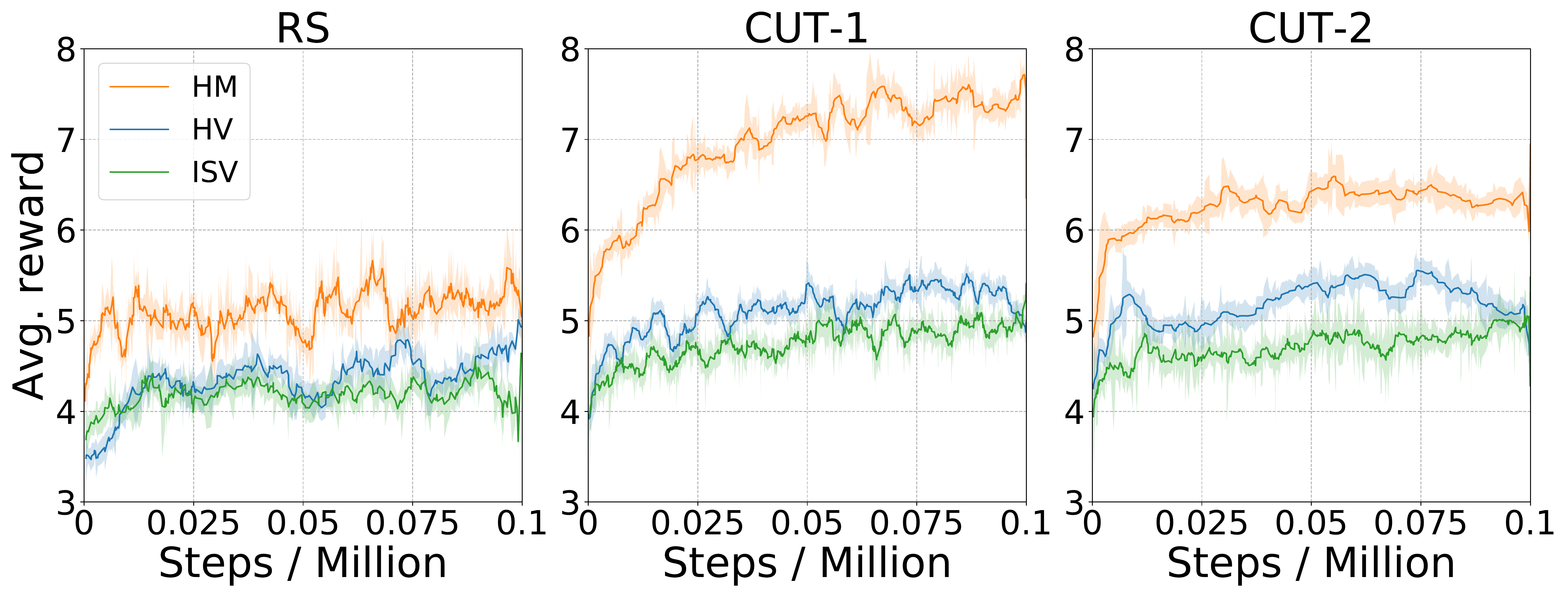} 
	\end{overpic}
    \caption{HM shows a clear advantage over vector-based height parameterizations (HV and ISV).}\label{fig:stateaction}\vspace{-4pt}
\end{figure}

\vspace{-10 pt}
\paragraph{Height parameterization}
Next, we show that the environment parameterization using the proposed 2D height map (HM) (i.e., the $\mathbf{H}$ matrix) is necessary and effective. To this end, we compare our method using HM against that employing two straightforward 1D alternatives. The first competitor is the height vector (HV), which is an $L \cdot W$-dimensional vector stacking columns of $\mathbf{H}$. The second competitor is referred to as the item sequence vector (ISV). The ISV lists all the information of items currently packed in the bin. Each packed item has 6 parameters corresponding to $X$, $Y$, and $Z$ coordinates of its FLB as well as the item's dimension.
From our test on CUT-1, HM leads to $16.0\%$ and $19.1\%$ higher space utilization and $4.3$ and $5.0$ more items packed than HV and ISV, respectively.
The plots in Figure~\ref{fig:stateaction} compare the average reward received using different parameterizations.
These results show that 2D height map (HM) is an effective way to describe the state-action space for 3D-BPP.

\begin{figure}[t!]\centering
    \begin{overpic}[width=1.0\linewidth,tics=10]{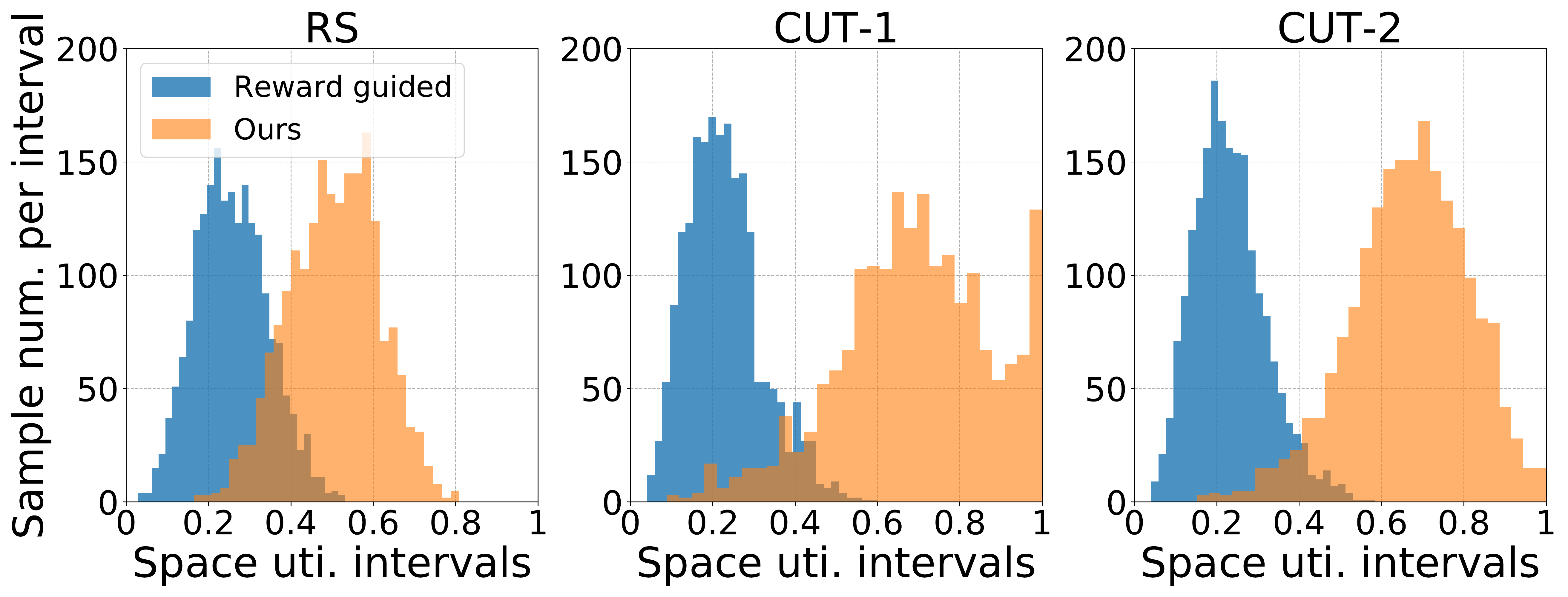}
    \end{overpic}
    \caption{Comparison to DRL with reward tuning. Our method obtains much better space utilization.}\label{fig:mask_penalty}\vspace{-16pt}
\end{figure}

\vspace{-10 pt}
\paragraph{Constraint vs. reward}
In DRL training, one usually discourages low-profile moves by tuning the reward function. We show that this strategy is less effective than our constraint-based method (i.e., learning invalid move by predicting the mask). In Figure~\ref{fig:mask_penalty}, we compare to an alternative method which uses a negative reward to penalize unsafe placements. Constraint-based DRL seldom predicts invalid moves (predicted placement are $99.5\%$ legit).

\begin{figure}[b]
	\begin{overpic}[width = \linewidth]{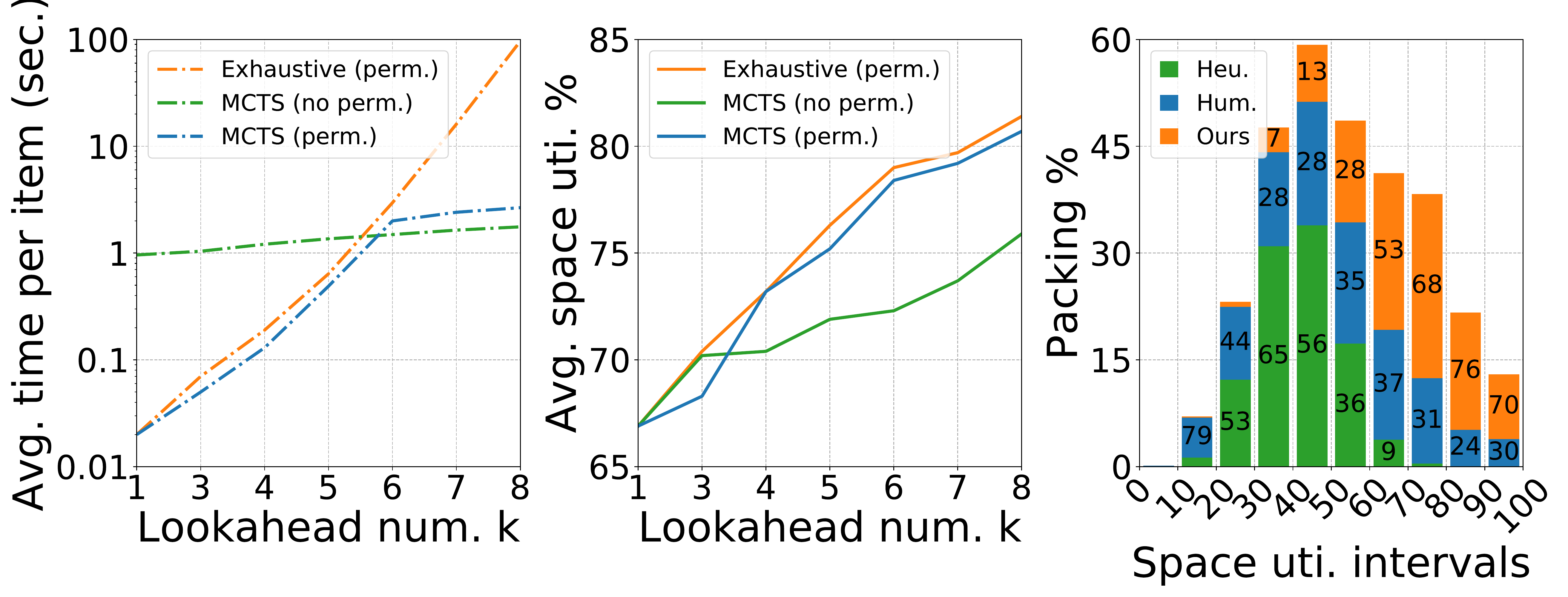}
    \put(19,37){\small (a)}
    \put(51,37){\small (b)}
    \put(83,37){\small (c)}
    \end{overpic}
    \vspace{-15 pt}
	\caption{(a): Our permutation based MCTS maintains good time efficiency as the number of lookahead items increases. (b): The performance of our MCTS based BPP-$k$ model achieves similar performance (avg. space utility) as the brute-force search over permutation tree. (c): The distribution of space utilization using boundary rule (Heu.), human intelligence (Hum.), and our BPP-$1$ method (Ours).}\label{fig:userstudy}
\end{figure}

\vspace{-10 pt}
\paragraph{Scalability of BPP-$k$}
With the capability of lookahead, it is expected that the agent better exploits the remaining space in the bin and delivers a more compact packing. On the other hand, due to the NP-hard nature, big $k$ values increase the environment space exponentially. Therefore, it is important to understand if MCTS is able to effectively navigate us in the space at the scale of $O(k!)$ for a good packing strategy. In Figure~\ref{fig:userstudy}(a,b), we compare our method with a brute-force permutation search, which traverses all $k!$ permutations of $k$ coming items and chooses the best packing strategy (i.e., the global optimal).
\kx{We also compare to MCTS-based action search with $k$ lookahead items in which no item permutation is involved.
We find that our MCTS-based permutation tree search yields the best results -- although having slightly lower space utilization rate ($\sim 3\%$), it is far more efficient.}
The search time of brute-force permutation quickly surpasses $100$s when $k=8$. Our method takes only $3.6$s even for $k=20$, when permutation needs hours. A larger $k$ makes the brute-force search computationally intractable.

\vspace{-10 pt}
\paragraph{Extension to different 3D-BPP variants}
Our method is versatile and can be easily generalized to handle different 3D-BPP variants such as admitting \emph{multiple bins} or allowing \emph{item re-orientation}. To realize multi-bin 3D-BPP, we initialize multiple BPP-$1$ instances matching the total bin number. When an item arrives, we pack it into the bin in which the item introduces the least drop of the critic value given by the corresponding BPP-$1$ network.
\supl{More details can be found in the supplemental material.}
Table~\ref{tab:multi-bin} shows our results for varying number of bins. More bins provide more options to host an item, thus leading to better performance (avg. number of items packed). Both time (decision time per item) and space complexities grow linearly with the number of bins.

\begin{table}
\vspace{-10 pt}
\begin{center}
	\scalebox{0.9}{
    \setlength{\tabcolsep}{1.2mm}{
	\begin{tabular}{c|c|c|c|c}
		\whline{1.15pt}
		{\# bins} & {Space uti.} & {\# items per bin} & {\# total items} & {Decision time} \\
		\whline{0.65pt}
		1 & $67.4\%$  & $17.6$     & $17.6$     & $2.2\times10^{-3}$ s \\
		4 & $69.4\%$  & $18.8$     & $75.2$     & $6.3\times10^{-3}$ s \\
		9 & $72.1\%$  & $19.1$     & $171.9$    & $1.8\times10^{-2}$ s \\
		16 & $75.3\%$  & $19.6$    & $313.6$    & $2.8\times10^{-2}$ s \\
		25 & $77.8\%$  & $20.2$    & $505.0$    & $4.5\times10^{-2}$ s \\
		\whline{1.15pt}
	\end{tabular}}}
\end{center}\vspace{-6pt}
\caption{Multi-bin packing tested with the CUT-2 dataset.}\label{tab:multi-bin}\vspace{-16pt}
\end{table}

We consider only horizontal, axis-align orientations of an item, which means that each item has two possible orientations. We therefore create two feasibility masks for each item, one for each orientation. The action space is also doubled. The network is then trained to output actions in the doubled action space. In our test on the RS dataset, we find allowing re-orientation increases the space utilization by $11.6\%$ and the average items packed by $3$, showing that our network handles well item re-orientation.

\begin{figure}
	\centering
	\includegraphics[width = 0.95\linewidth]{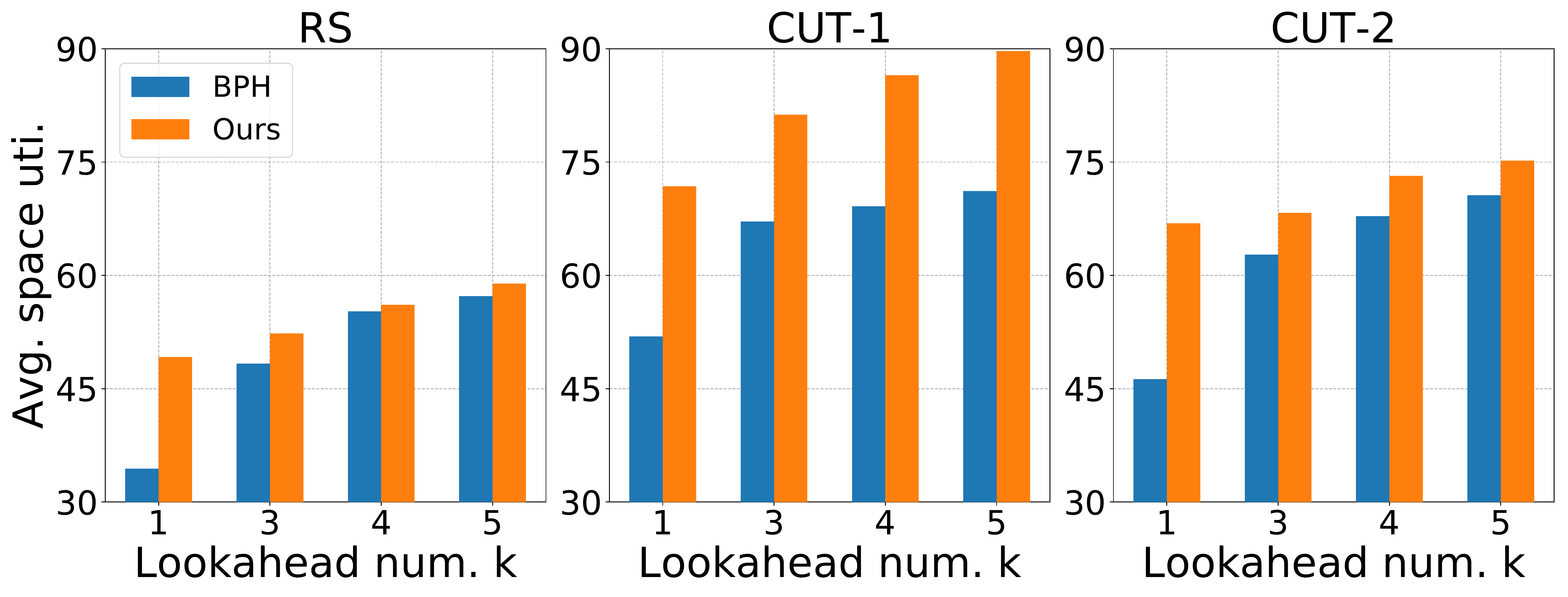} 
	\caption{\kx{Comparison with the online BPH method~\cite{ha2017online} on BPP-$k$. Note that BPH allows lookahead item re-ordering while ours does not. 
	}}\label{fig:bppkcomparison}\vspace{-14pt}
\end{figure}

\vspace{-10 pt}
\paragraph{Comparison with non-learning methods}
\kx{
Existing works mostly study offline BPP and usually adopt non-learning methods.
We compare to two representatives with source code available.
The first is a heuristic-based online approach, BPH~\cite{ha2017online} which allows the agent to select the next best item from $k$ lookahead ones (i.e., BPP-$k$ with re-ordering). In Table~\ref{tab:comparison}, we compare to its BPP-$1$ version to be fair. In Figure~\ref{fig:bppkcomparison}, we compare \emph{online BPH} and our method under the setting of BPP-$k$.
Most existing methods focus on offline packing where the full sequence of items is known \emph{a priori}.
The second method is the \emph{offline LBP} method~\cite{martello2000three} which is again heuristic based.
In addition, we also design a heuristic baseline which we call \emph{boundary rule} method.
It replicates human's behavior by trying to place a new item side-by-side with the existing packed items and keep the packing volume as regular as possible (\supl{details in the supplemental material}).

From the comparison in Table~\ref{tab:comparison}, our method outperforms all alternative online methods on all three benchmarks and even beats the offline approach on CUT-1 and CUT-2. Through examining the packing results visually, we find that our method automatically learns the above ``boundary rule'' even without imposing such constraints explicitly.
From Figure~\ref{fig:bppkcomparison}, our method performs better than \emph{online BPH} consistently with varying number of lookahead items even though BPH allows re-ordering of the lookahead items.

We also conducted a preliminary comparison on a real robot test of BPP-1 (see our accompanying video). Over $50$ random item sequences, our method achieves averagely $66.3\%$ space utilization, much higher than \emph{boundary rule} ($39.2\%$) and \emph{online BPH} ($43.2\%$).
}


\input{comparison}

\vspace{-10pt}
\paragraph{Our method vs. human intelligence}
The strongest competitor to all heuristic algorithms may be human intuition.
To this end, we created a simple Sokoban-like app (\supl{see the supplemental material}) and asked $50$ human users to pack items manually vs. AI (our method). The winner is the one with a higher space utilization rate. $15$ of the users are palletizing workers and the rest are CS-majored undergraduate/graduate students. We do not impose any time limits to the user. The statistics are plotted in Figure~\ref{fig:userstudy}(c). To our surprise, our method outperforms human players in general ($1,339$ AI wins vs. $406$ human wins and $98$ evens): it achieves $68.9\%$ average space utilization over $1,851$ games, while human players only have $52.1\%$.

%% file: comparison.tex
\begin{table}[t!]\centering
\vspace{-6pt}
\scalebox{0.8}{
\begin{tabular}{c|ccc}
	\whline{1.15pt}
	\multirow{2}{*}{Method} & \multicolumn{3}{c}{\# items / \% Space uti.}\\
	& RS          & CUT-1       & CUT-2   \\
	\whline{1.15pt}
	Boundary rule (Online) & 8.7 / 34.9\%      & 10.8 / 41.2\%     & 11.1 / 40.8\% \\
	BPH (Online) & 8.7 / 35.4\% & 13.5 / 51.9\%      & 13.1 / 49.2\% \\
    LBP (Offline) & \textbf{12.9} / \textbf{54.7\%} & 14.9 / 59.1\%      & 15.2 / 59.5\% \\
    Our BPP-$1$ (Online)           & 12.2 / 50.5\%      & \textbf{19.1} / \textbf{73.4\%}      & \textbf{17.5} / \textbf{66.9\%}   \\
	\whline{1.15pt}
\end{tabular}
}\vspace{-3pt}
\caption{Comparison with three baselines including both online and offline approaches.}\label{tab:comparison}
\vspace{-12pt}
\end{table}

%% file: conclusion.tex
\section{Conclusion}
\label{sec:conclusion}
We have tackled a challenging online 3D-BPP via formulating it as a constrained Markov decision process and solving it with constrained DRL. The constraints include order dependence and physical stability. Within the actor-critic framework, we achieve policy optimization subject to the complicated constraints based on a height-map bin representation and action feasibility prediction.
In realizing BPP with multiple lookahead items,
we adopt MCTS to search the best action over different permutations of the lookahead items.
In the future, we would like to investigate more relaxations of the problem. For example, one could lift the order dependence constraint by adding a buffer zone smaller than $|\mathcal{I}_o|$.
\kx{Another more challenging relaxation is to learn to pack items with irregular shape.}

%% file: supp.tex
\section{Supplemental material overview}\label{ssec:intro}
In this supplemental document, we report more implementation and experiment
details. 

\begin{itemize}
	\item Section~\ref{sec:implementation} gives more descriptions
	regarding the network architecture, training selection, Monte Carlo permutation
	tree search, etc.
	\item Section~\ref{sec:benchmark} elaborates how CUT-1, CUT-2, and RS are constructed. Details about the heuristic baseline we compared in our experiment.
	\item The user study design is reported in Section~\ref{sec:userstudy}.
	\item Section~\ref{sec:step} analyzes 
	the performance difference between step-wise reward and termination reward in our problem.
	\item The details of reward function to penalize unsafe placements is reported in Section~\ref{sec:penalization}
	\item More experiment results are reported in Section~\ref{sec:moreresults}. 
\end{itemize}


\section{Implementation Details}\label{sec:implementation}
We report the details of our implementation in this section, and our source
code is also submitted with this supplemental material.

\begin{figure}
	\centering
	\includegraphics[width=0.96\linewidth]{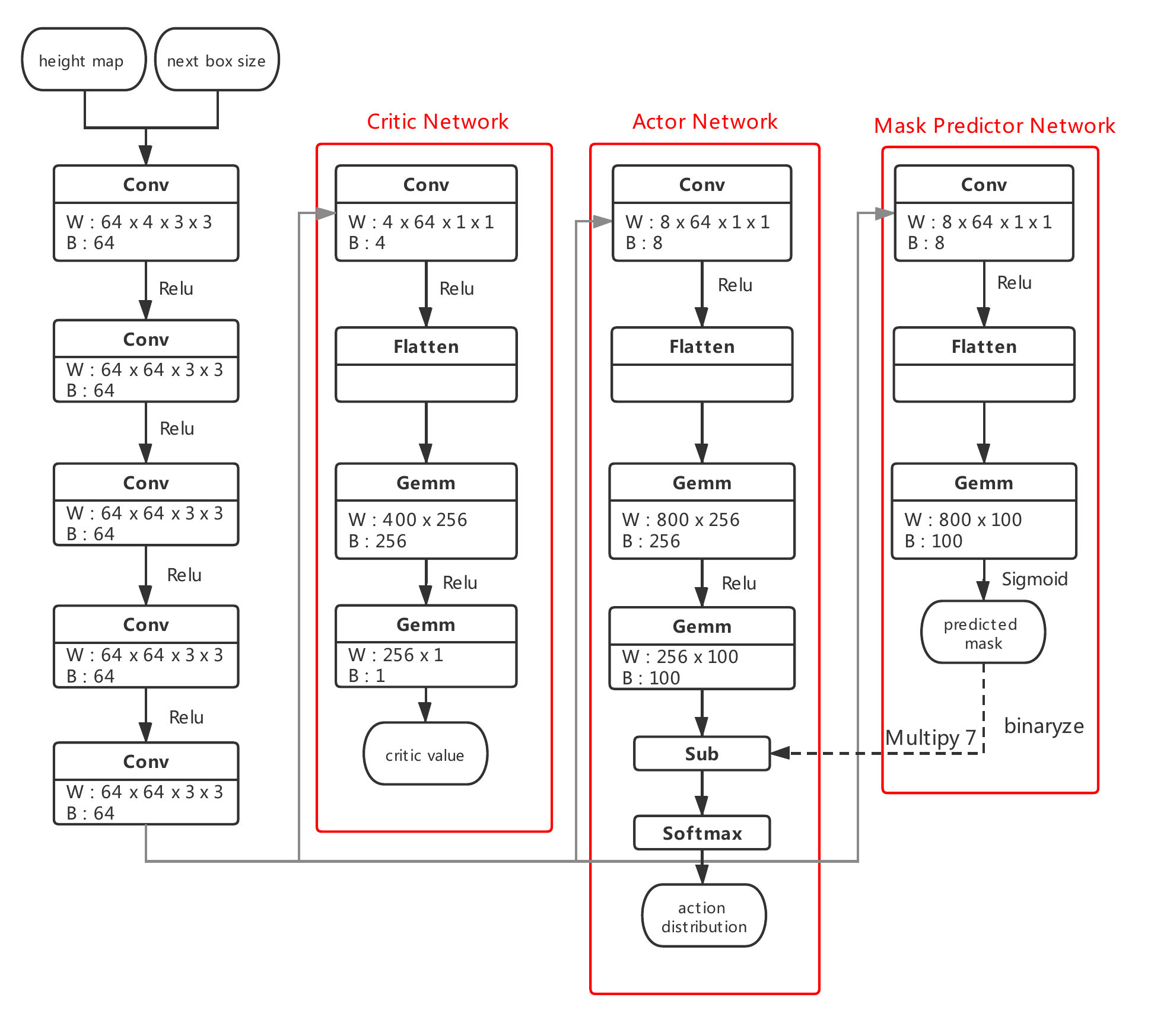}
	\caption{Detailed network architecture.}
	\label{fig:network}
\end{figure}

\paragraph{Network architecture and training configurations}
A detailed specifications of the our network is shown in Figure~\ref{fig:network}. Our pipeline consists of three major
components: an actor network, a critic network, and the feasibility mask
predictor. It takes three inputs, \emph{height map} $\mathbf{H}_n$ and the
dimensionality  $\mathbf{d}_n = [l_n, w_n, h_n]^\top\in\mathbb{Z}^3$ of the
current item $n$ to be packed as state, and the feasibility mask
$\mathbf{M}_n$ as ground truth. Note that $\mathbf{M}_n$ is only used in the training
processing.

The whole network is trained via a composite loss consisting of actor loss
$L_{actor}$, critic loss $L_{critic}$, mask prediction loss $L_{mask}$,
infeasibility loss $E_{inf}$ and action entropy loss $E_{entropy}$. These loss
function are defined as:

\begin{equation}
\left\{
\begin{array}{ll}
	\displaystyle L_{actor} & \displaystyle =  (\mathbf{R}_n-V(s_n))\log
P(a_n | s_n) \\
	\displaystyle L_{critic} & \displaystyle = (\mathbf{R}_n-V(s_n))^2  \\
	\displaystyle L_{mask} &
\displaystyle =\sum_{(x,y)}(\mathbf{M}_n^{gt}-\mathbf{M}_n^{pred})^2\\
\displaystyle {E_{inf}} &\displaystyle =\sum_{{\mathbf{M}_n(x,y)=0}} P(a_n=L \cdot x + y|s_n)\\
\displaystyle E_{entropy} & \displaystyle= \sum_{\mathbf{M}_n(x,y)=1}-P(a_n |s_n) \cdot
\log \big(P(a_n |s_n)\big),
\end{array}
\right.
\end{equation}	

where  $\displaystyle \mathbf{R}_n=r_{n}+
	\gamma  V(s_{n+1})$ and
$\displaystyle r_n = 
10 \times l_n \cdot w_n \cdot h_n/(L \cdot W \cdot H) $ is our reward
function which indicates the space utilization. When the current item is not placeable, its reward is zero and the packing sequence ends. 
Here, $\gamma \in [0,1]$ is the discount factor and we set $\gamma$ as 1 so that $\displaystyle \mathbf{R}_n$ can directly present how much utilization can agent obtain from $\displaystyle s_n$ on.
The output of critic network $\displaystyle V(s_{n})$ would give a state value prediction of $\displaystyle s_n$ and help the training of actor network which outputs a possibility matrix of the next move. This probability is scaled based on $\mathbf{M}_n$ --- if the move is infeasible, the possibility will be multiplied by a penalty factor of $0.001$. Afterwards, a softmax operation is adopted to output the final action distribution.
Note that, the infeasibility penalty could be absent in test with the help of $E_{inf}$ and our method can still work well in this situation.

\begin{figure*}[!t]
	\centering
	\subfigure[Packing performance on RS.]{
	\begin{minipage}[t]{0.33\linewidth}
	\centering
	\includegraphics[width=0.93\linewidth]{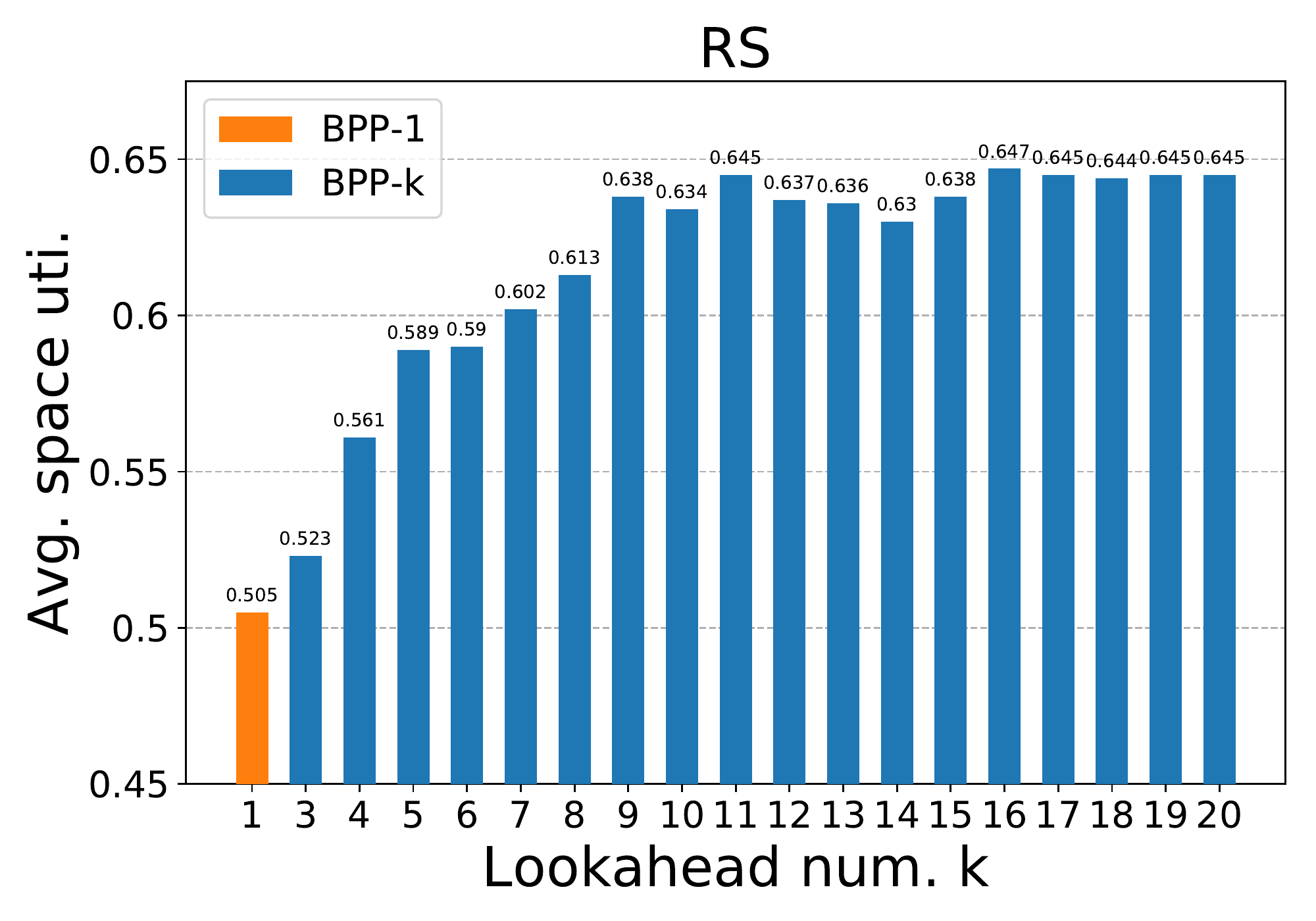}
	\end{minipage}%
	}%
	\subfigure[Packing performance on CUT-1.]{
	\begin{minipage}[t]{0.33\linewidth}
	\centering
	\includegraphics[width=0.93\linewidth]{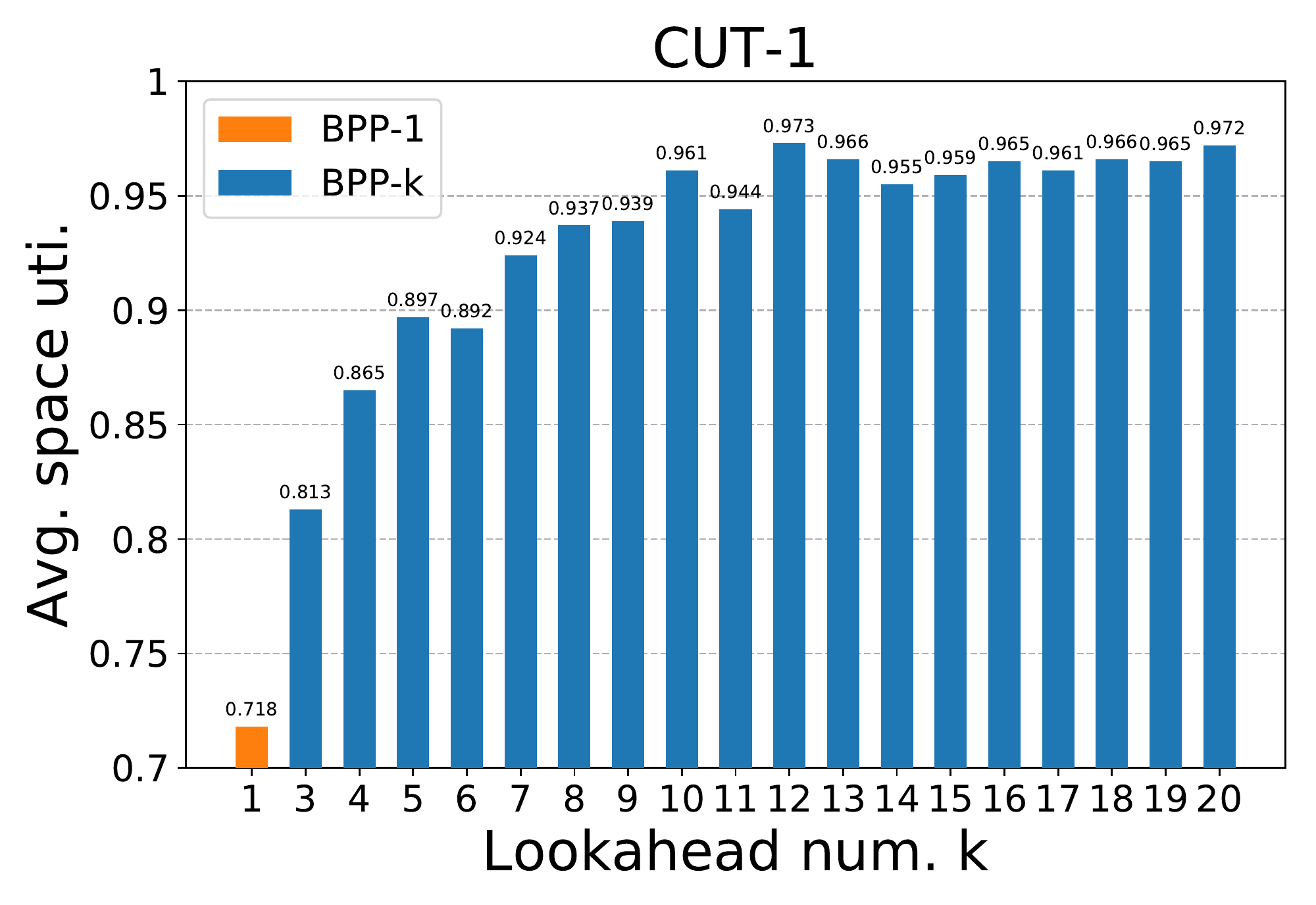}
	\end{minipage}%
	}%
	\subfigure[Packing performance on CUT-2.]{
	\begin{minipage}[t]{0.33\linewidth}
	\centering
	\includegraphics[width=0.93\linewidth]{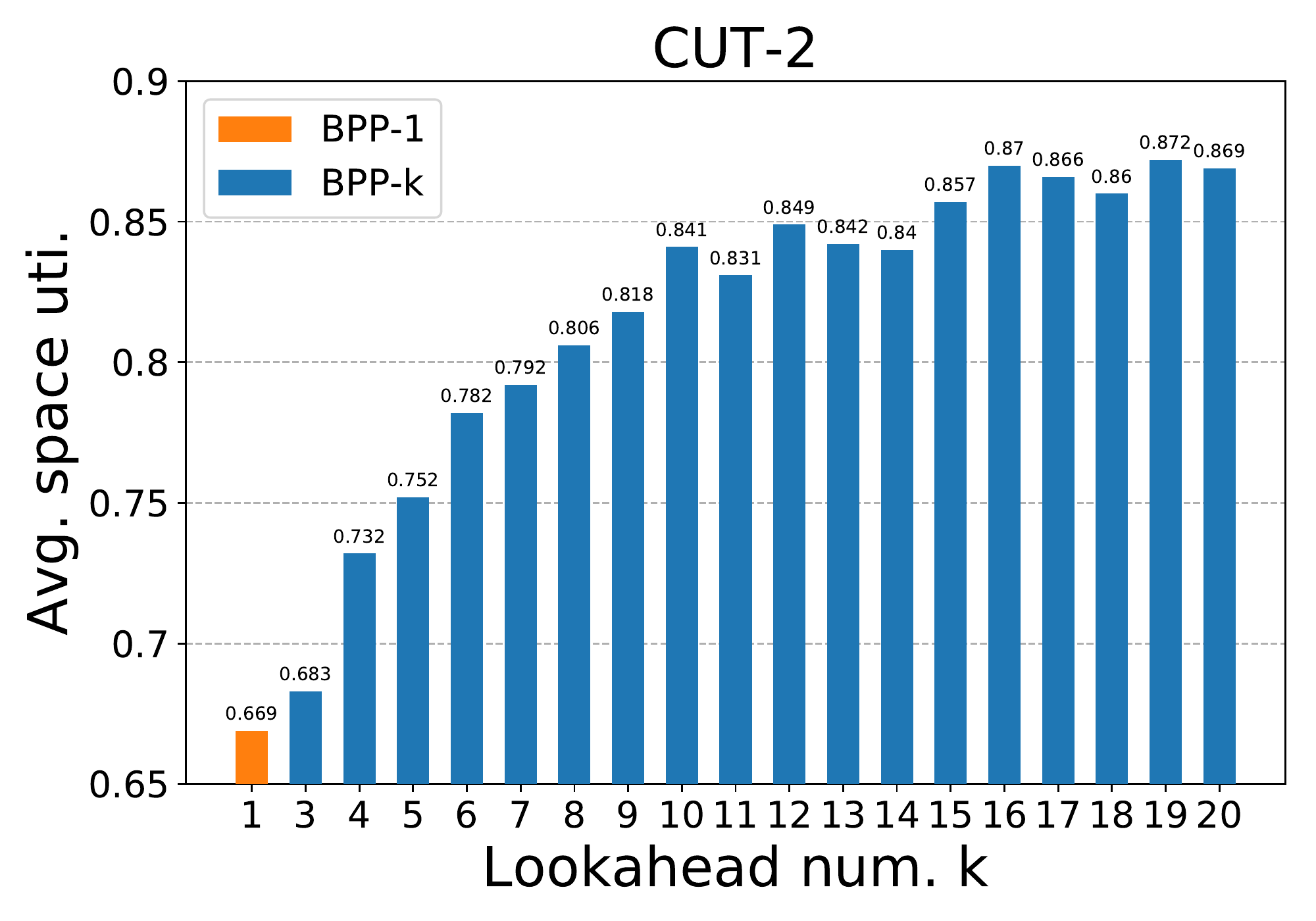}
	\end{minipage}
	}%
	\centering
	\caption{When $k$ increases, the space utilization rate first goes up and later, enters a ``plateau'' zone.}\label{fig:mcts}
\end{figure*}

\paragraph{Monte Carlo permutation tree search}
Our algorithm is inspired by the Monte Carlo tree search of~\cite{silver2017mastering}. The main difference lies in: Firstly, the goal of our MCTS is to find the best packing order for the next $k$ items; Secondly, max reward is used in our MCTS instead of the mean reward as the evaluation strategy. Algorithm~\ref{alg:mcts} outlines the entire procedure step by step, where in $T$ is the maximum simulation time, $I$ is lookahead items, $Last$ is the first item after I, $i_0$ is current item, $s$ is state input of each node and $a$ is action of each
node. Environment simulator $SimEnv$, which takes height map, the next item dimension, and action (position of item) as the input, and returns the updated height map. Action choosing function $\pi$ uses policy network from BPP-1 model to get the action with highest possibility. N is visit times of a node. Q is expect return from a node. We test our method on our three benchmarks, and more results can be found in Figure~\ref{fig:mcts}. 



\paragraph{Multi-bin scoring}
It is straightforward to generalize our method for multiple bins. The only difference is that the agent needs to determine which bin the item is to be packed and the rest is naturally reduced to a single-bin BPP instance. 
To this end, we estimate the packing score for all the available bins and pack the item into the one with highest scores.
This score of each bin indicates how much the state value changes if the current item is packed into it. 
The state value is given by the value network, which is a estimation value of how much reward will we get from selected bin. If this value of the selected bin drops significantly while an item is packed in it, it implies that we should pack this item into other bins.
Our bin selection method is described in Algorithm~\ref{alg:multibin}. 
Here, $val$ is the last state value estimation of bin $b$, $V$ is the state value estimation function via a 
value network, $n$ is the current item, $B$ is the set of bins and $H$ is the height maps of bins. 
The default score for each bin at beginning $s_{def}=-0.2$.

			
			

\paragraph{Orientated items}
Our method can also incorporating items with different orientation. We multifold the action space and related mask based on how many different orientations are considered, e.g. we will have a $m$ times larger feasibility $M_n$ and action space if $m$ different poses are allowed for an item. Doing so induces more flexibility for the packing, and it potentially leads to a better result. This is observed and reported in Table~\ref{tab:orientation}. Note that, orientation only happens around Z axis in our problem setting.

\begin{table}[h]
	\centering
	\caption{Performance comparison with and without orientation on different benchmarks.}\label{tab:orientation}
	\scalebox{0.99}{
 		\begin{tabular}{c|c|c|c}
			\whline{1.15pt}
 			{} & {RS} & {CUT-1} & {CUT-2} \\
 			\whline{0.65pt}
 			w orientation & $62.1\%$  & $76.2\%$ & $70.2\%$    \\
 			w/o orientation & $50.5\%$  & $73.4\%$  & $66.9\%$       \\			
 			\whline{1.15pt}
			\end{tabular}}\\
\end{table}

\section{Benchmark Construction}\label{sec:benchmark}
\begin{figure*}[ht!]
	\centering
	\includegraphics[width=\linewidth]{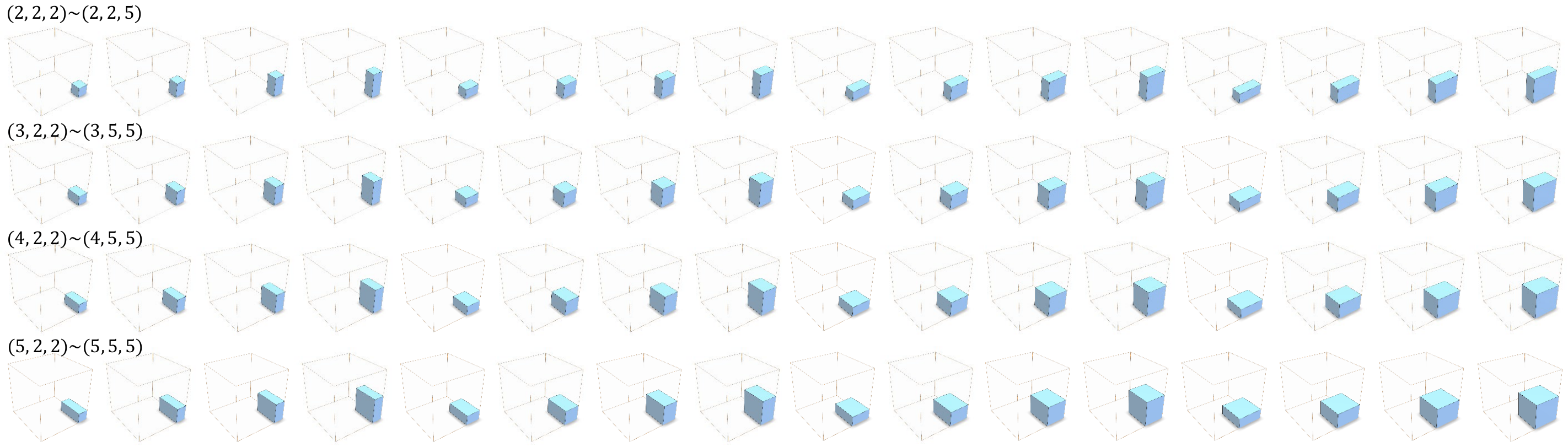}
	\caption{Pre-defined item set $\mathcal{I}$.}
	\label{fig:boxes}
\end{figure*}

All 64 pre-defined items are visualized in Figure~\ref{fig:boxes}. Algorithm~\ref{alg:construction} outlines how the dataset is constructed  given the bin size $L$, $W$, $H$ and a valid item size threshold.

The sequence in RS benchmark is generated by random sampling. Each item
along the sequence is picked out of our pre-defined item set $\mathcal{I}$ randomly. However, as everything is random, we do not know the optimal packing configuration of a RS sequence ourselves (unless we run an exhaustive branch-and-bound search~\cite{martello2000three} which is much too time consuming to accomplish). For a better quantitative evaluation, we also generate item sequences via cutting stock~\cite{gilmore1961linear}. It is clear that a sequence created by cutting the bin should be packed in bin perfectly with a perfect space utilization of $100\%$. Algorithm~\ref{alg:construction} provides the detailed procedures of the data generation.

\begin{figure}[h]
	\centering
	\includegraphics[width=0.9\linewidth]{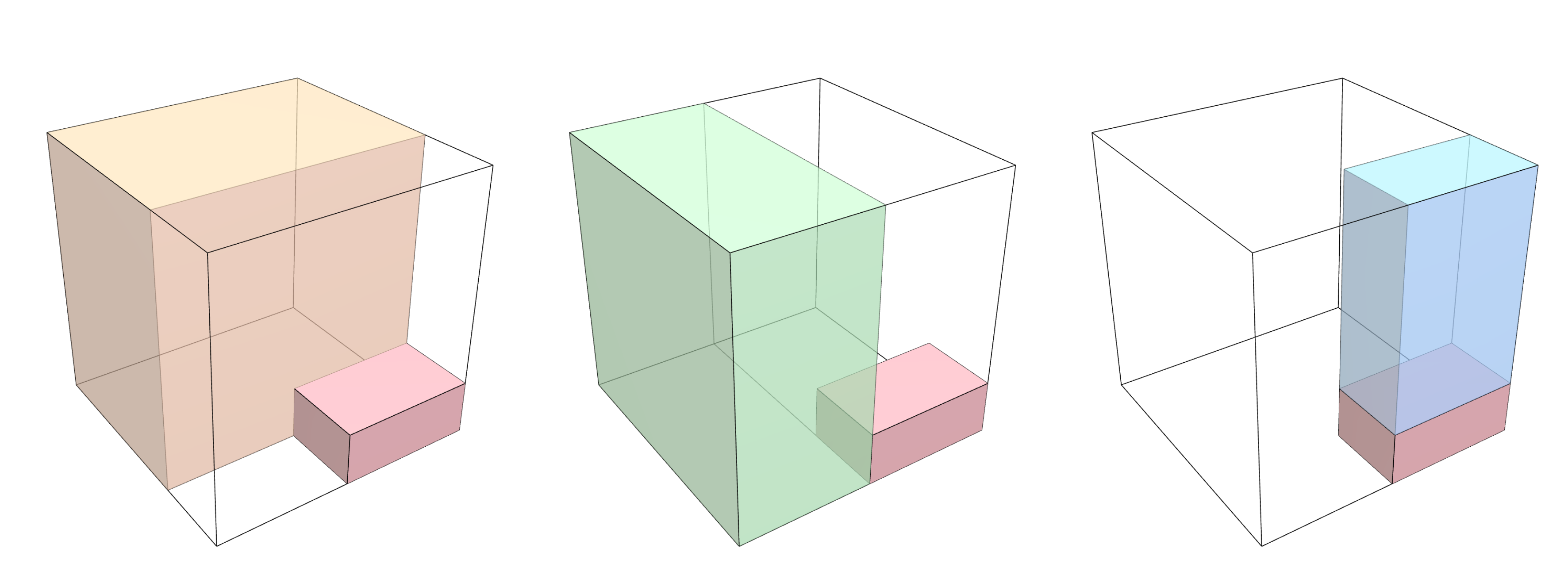}
	\caption{The maximum spare cuboid.}
	\label{fig:gap}
\end{figure}

\section{Heuristic Baseline Method}\label{sec:basline}
Online BPP is an under-investigated problem. To better demonstrate the effectiveness of our method, we design
 a heuristic baseline approach to evaluate the performance of our DRL method. We report details of this baseline approach in this section. This method is designed based on a simple observation, human would try to keep the volume of packed bin to be regular during the packing to maintain the left regular space as large as possible. Such ``regularity'' is used as the metric to measure a packing action.

To describe regularity of a bin, we introduce the concept of \emph{spare cuboid}. As shown in Figure~\ref{fig:gap}, a spare cuboid is an unoccupied, rectangular space in the bin, and the regularity of a bin is defined based on the \emph{maximum} spare cuboids. Intuitively, we would like to have a bigger maximum spare cuboid. If a packed bin has many small-size spare cuboids, it implies the remaining space of this bin is not ``regular''. 
As illustrated in Figure~\ref{fig:gap}, the right packing strategy would left the biggest spare cuboid. 
The regularity of the bin is then defined as the maximum rectangular residual space or \emph{maximum spare cuboid}. Since $\mathcal{I}$ is pre-defined, we know how many items can be packed into a maximum spare cuboid. Based on this, we rate each maximum spare cuboid $c$ by the number of item types can be packed in $RS_c=\|\mathcal{I}_{valid}\|+c_{volume},\mathcal{I}_{valid} \subset \mathcal{I}$. If a maximum spare cuboid fits all the items in $\mathcal{I}$, additional reward is given as: $RS_c=\|\mathcal{I}\|+c_{volume}+10$. The final score $BS_p$ of a bin by packing the current item at $p$ would be the sum of $RS_c$ of its maximum spare cuboid $c$. And we can find the best packing position $p_{best}$ as:

\begin{equation}
p_{best} = \arg \max_p \frac{1}{\|\mathcal{C}\|} \sum_{c \in \mathcal{C}} RS_c
\end{equation}

\begin{figure}[h]
	\centering
	\includegraphics[width=0.86\linewidth]{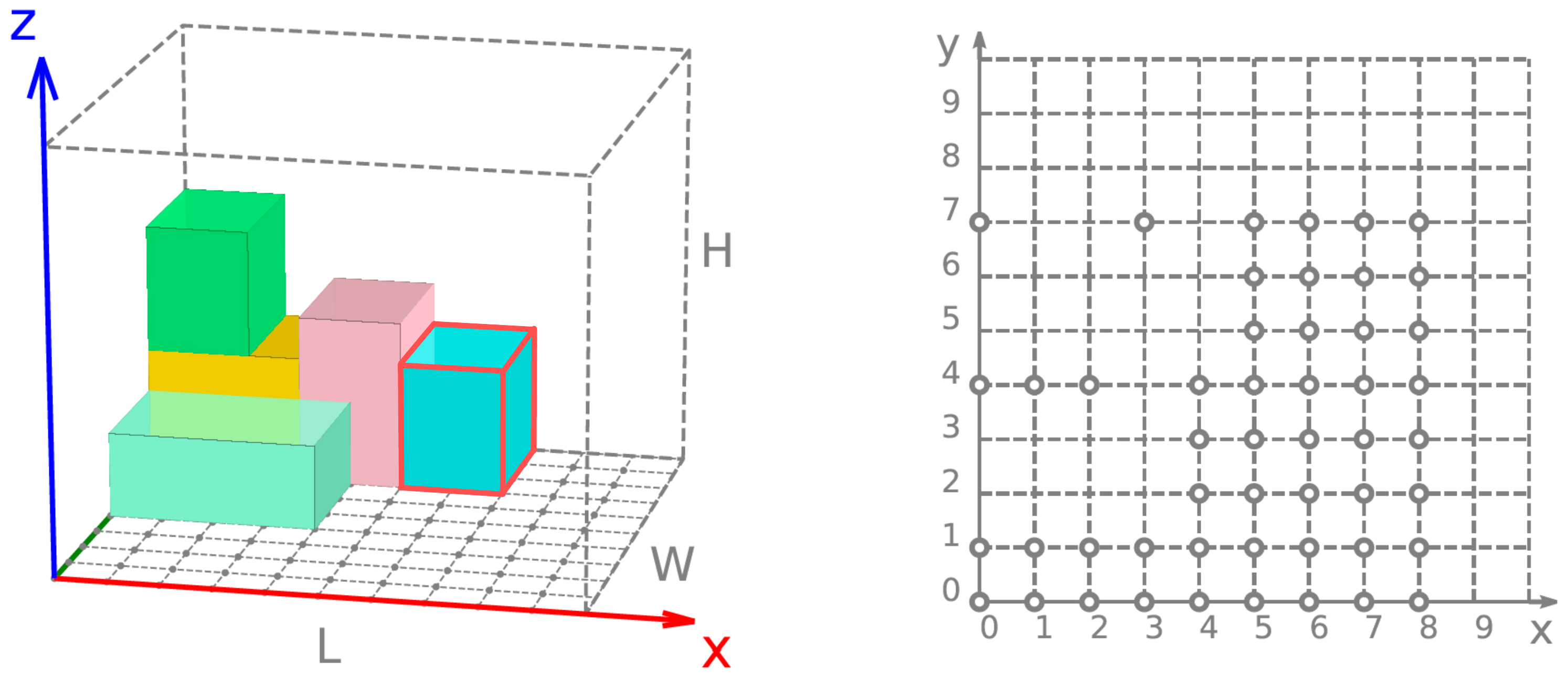}
	\caption{Left: The 3D visualization of packed items in our user study app. The current item is highlighted with red frame. Right: The action space of the bin --- the empty circles indicate suggested placements for the current item.}
	\label{fig:user}
\end{figure}

\begin{figure*}[ht!]
	\centering
\begin{minipage}{0.6\columnwidth}
	\includegraphics[width=\linewidth]{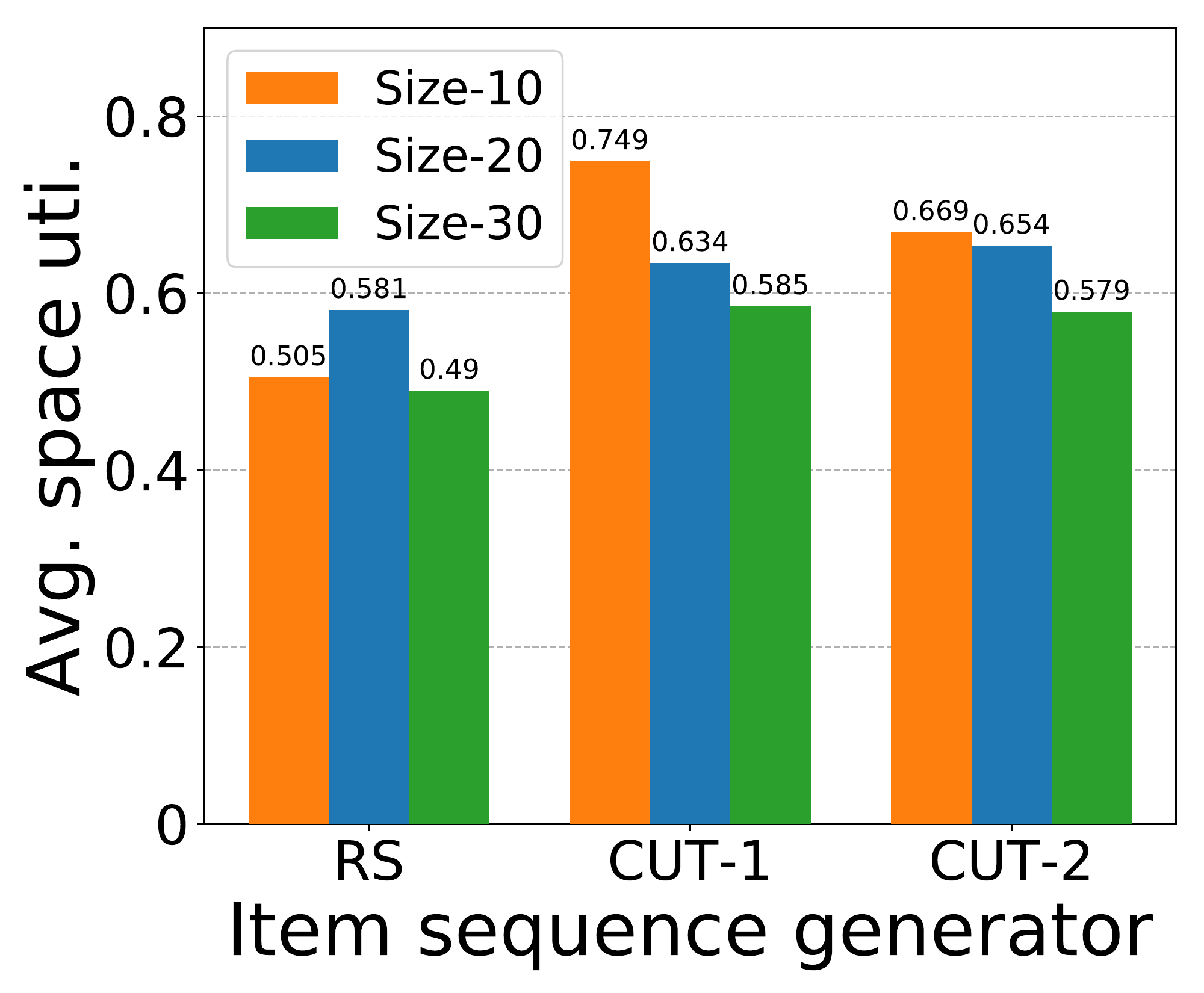}
	\centering
\end{minipage}\hspace{10pt}
\begin{minipage}{1.3\columnwidth}
	\includegraphics[width=\linewidth]{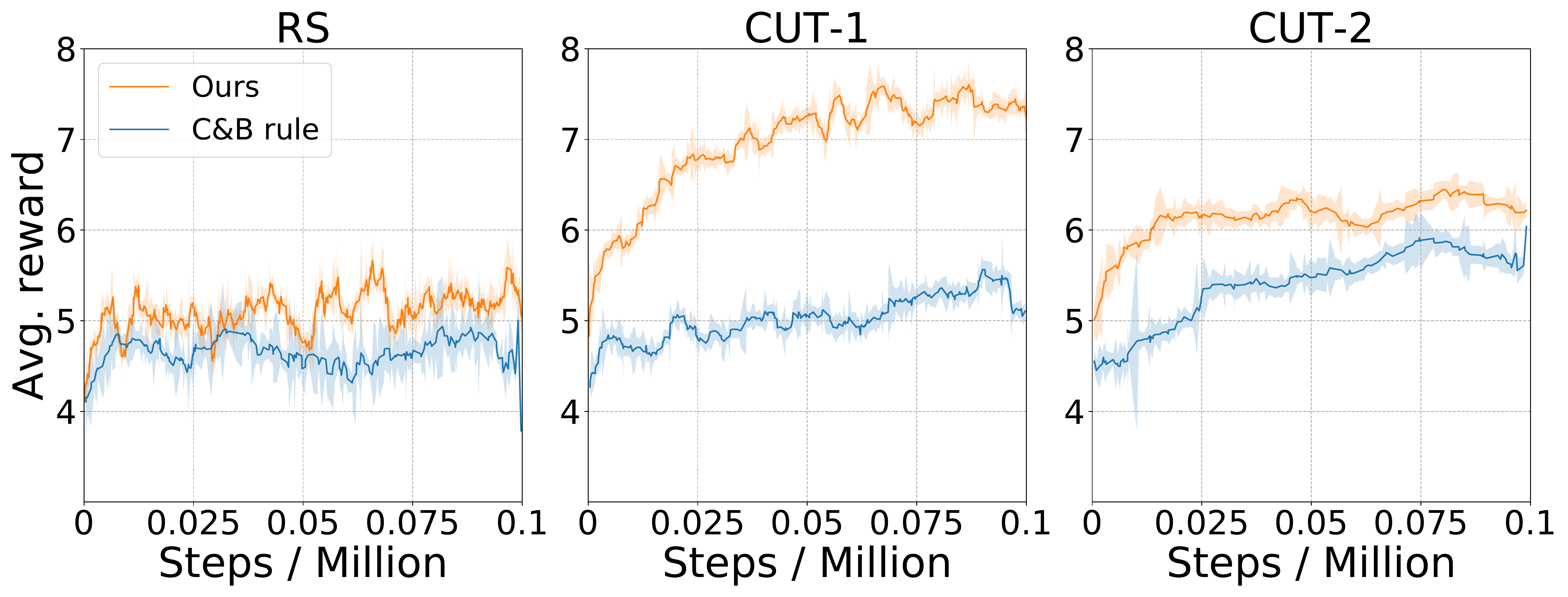}
	\centering
\end{minipage}
\caption{Left: Packing performance on different resolution. Size-10
means the resolution of the test bin is $10\times10\times10$ and etc. Second to right: Imposing the C\&B rule leads to inferior
performance (lower average reward).}\label{fig:scale}
\end{figure*}

\section{User Study}\label{sec:userstudy}

Figure~\ref{fig:user} is the interface of our user study app, which consists of two parts: visualization and action space. The test sequences is randomly picked from CUT-2 test set. Users can drag our UI to change the angle of view thus having a full observation of the packed items. To help users make better decision, our app allow them choose any suggestion circle in action space and virtually place item before they make the final decision. No time limit is given. When there is no suitable place for the current item, the test will reset and the selected sequence will be saved.

\section{Reward function design}\label{sec:step}
In this section we analyze the design of our reward function. We have two candidate reward functions, step-wise reward and termination reward.
For current item $n$, the step-wise reward is defined as $\displaystyle r_n = 10 \times l_n \cdot w_n \cdot h_n/(L \cdot W \cdot H) $ if $n$ is placed successfully otherwise $r_n = 0$. 
Meanwhile the termination reward is defined as the final capacity utilization $r = \sum_i^{i<n} 10 \times l_i \cdot w_i \cdot h_i/(L \cdot W \cdot H) \in (0,10]$, and it is only functional if the packing of a sequence is terminated.
We evaluate the performance of these two reward functions and the result is presented in Table~\ref{tab:step}. 

\begin{table}[h]
	\centering
	\caption{Extra information enables agent make use of termination reward.}\label{tab:step}
	\scalebox{0.99}{
 		\begin{tabular}{c|c|c|c}
			\whline{1.15pt}
 			{} & {RS} & {CUT-1} & {CUT-2} \\
 			\whline{0.65pt}
 			step-wise reward       & $50.5\%$  & $73.4\%$  & $66.9\%$    \\
			termination reward 		   & $39.2\%$  & $72.5\%$  & $66.5\%$       \\			
			termination reward \& uti.  & $50.3\%$  & $73.2\%$  & $66.7\%$       \\			
 			\whline{1.15pt}
			\end{tabular}}\\
\end{table}

\begin{figure}[h]
	\centering
	\includegraphics[width=\linewidth]{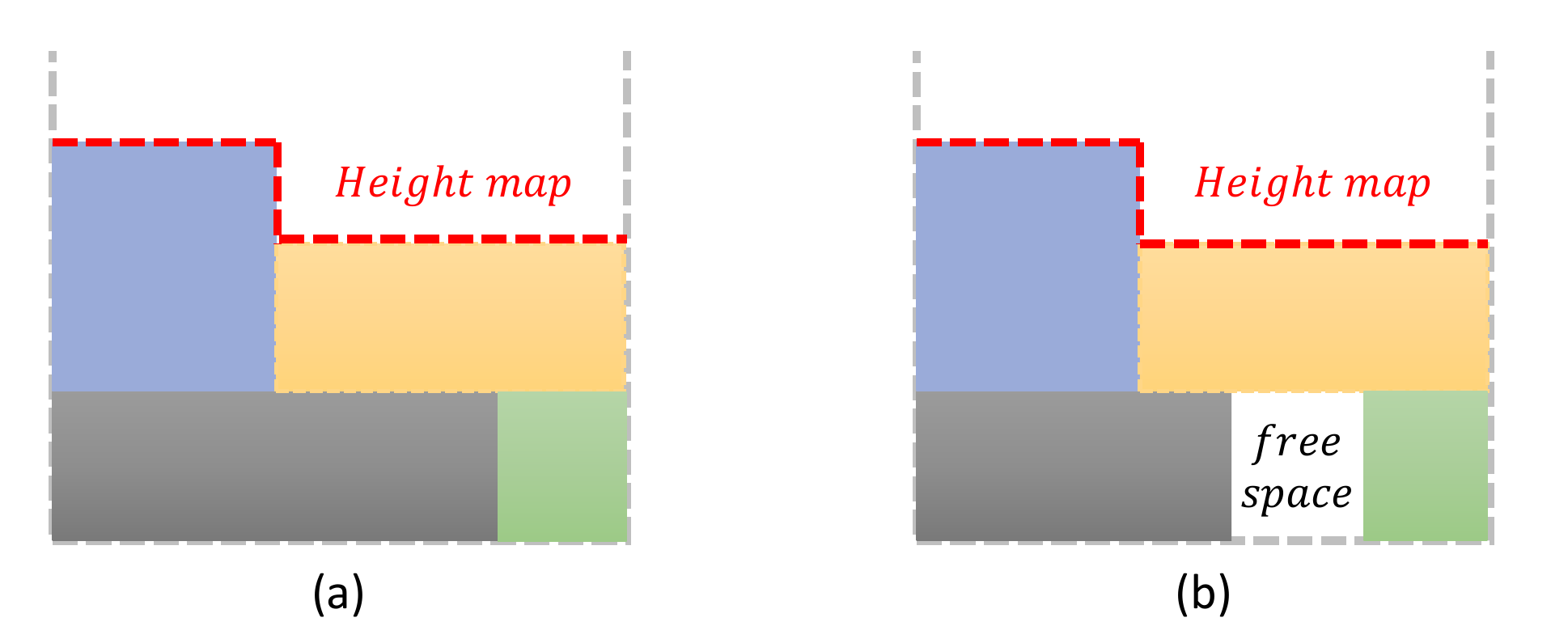}
	\caption{Both (a) and (b) has same height map but different space utilization, which is ambiguous for agent to predict state value given termination reward.}

	\label{fig:empty}
\end{figure}

The termination reward can perform similar with step-wise reward on CUT-1 and CUT-2 benchmarks. However, it doesn't perform well on RS benchmark due to the construction approach of sequences in RS can not guarantee the height map is enough for describing the packed bin as illustrated in Figure~\ref{fig:empty}. In other cases, the step-wise reward which focuses more on how much space is left above the packed items at each step, it makes this reward function can perform well even on RS benchmark.

We also design an experiment to further investigate the above assumption about these two reward functions. We encode an additional matrix as input for the termination reward which indicates whether there exists free space below height map. In this case, the state input would no longer be ambiguous for agent to perform prediction.
Table~\ref{tab:step} demonstrates that with additional information, performance on termination reward nearly equals to step-wise one. While CUT-1 and CUT-2 can be packed relatively close and less free space under height map exists, termination reward doesn't 
affect performance of these benchmarks too much.


\section{Penalized reward}\label{sec:penalization}

To explore whether reward guided based DRL or our constraint based DRL can help the agent avoiding to place item in unsafe place better, we design a reward guided based DRL alternative for comparison in our main paper. This section will report this reward guided based DRL alternative detailedly.

The basic idea to design this guided based DRL method is reward the agent when it packing item in a safe place and penalize it when it perform a dangerous move. The reward is designed as below: If item $n$ is placed successfully, the agent would be awarded as $r_n = 10 \times l_n \cdot w_n \cdot h_n / (L \cdot W \cdot H )$. Otherwise, if the placement of item $n$ violates the physical stability, the agent would be penalized as $r_n = -1$ and the packing sequence would be terminated.

We found even we explicitly penalize the agent when placing items on unsafe places during the training, the agent 
will still make mistakes every now and then in the test. Reward shaping cannot guarantee the placement safety as our constraint based DRL method.

\section{More Results}\label{sec:moreresults}

\begin{figure*}[b]
	\centering
	\includegraphics[width=\linewidth]{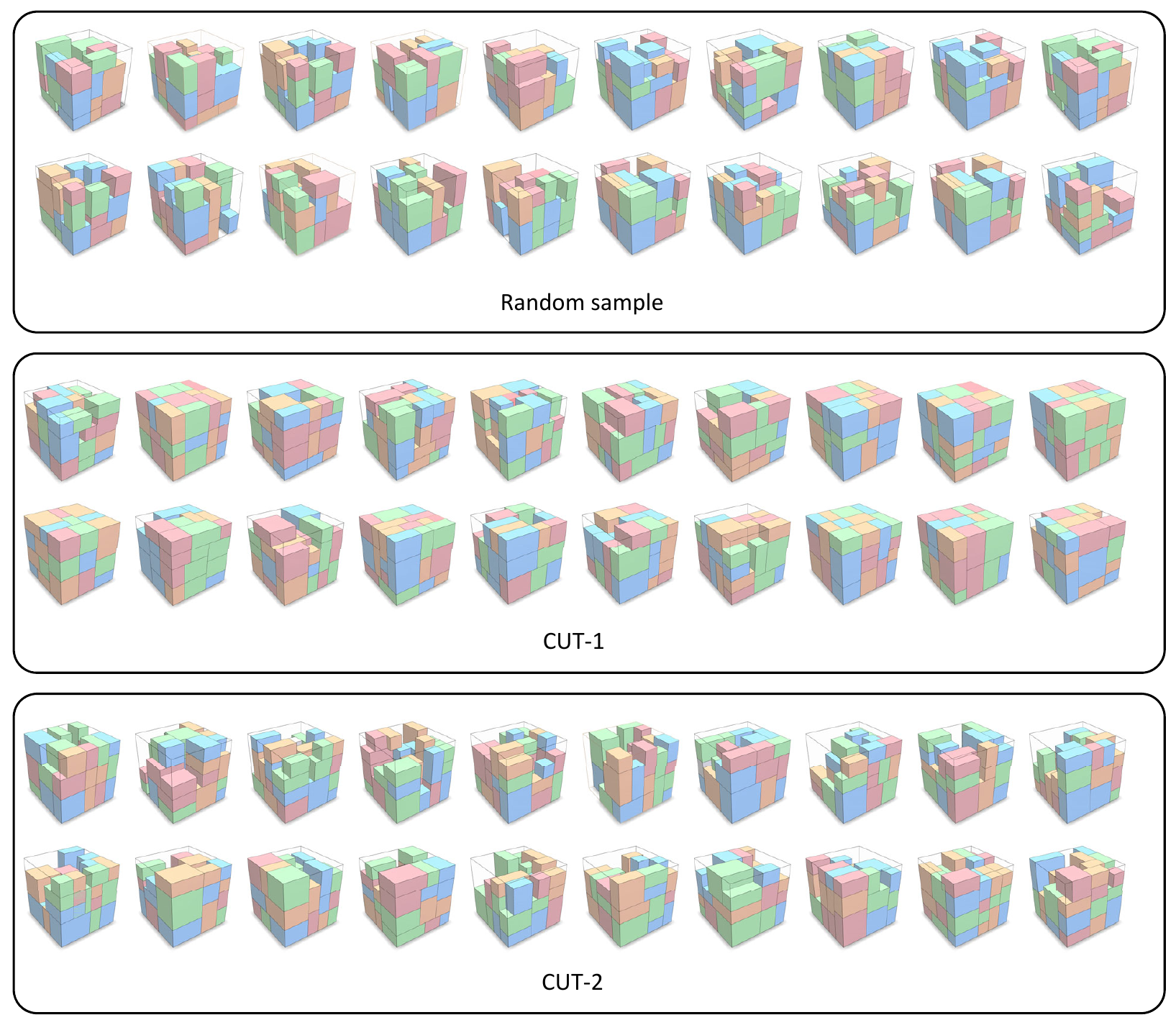}
	\caption{Packing results of our BPP-$1$ model.}
	\label{fig:results}
\end{figure*}

\paragraph{More visualized results.} Figure~\ref{fig:results} shows more packing results on three different benchmarks. An animated packing can be found in the supplemental video.

\paragraph{Study of action space resolution.} We also investigate how the resolution of the bin would affect the our performance. In this experiment, we increase the spatial discretization from $10\times10\times10$ to $20\times20\times20$ and $30\times30\times30$. As shown in Figure~\ref{fig:scale}, the performance only slightly decrease. Increased discretization widens the distribution of possible action space and dilutes the weight of the optimal action. However, our method remains efficient even when the problem complexity is $\sim 27\times$ bigger. This experiment demonstrates a good scalability of our method in a high-resolution environment.

\paragraph{Learned vs. heuristic strategy}
In real-world bin packing, human tends to place a box to touch the sides or corners of the bin or the other already placed boxes. We refer to this intuitive strategy as \emph{corner \& boundary rule} (C\&B rule). An interesting discovery from our experiment is that our method can automatically learn when to follow the C\&B rule smartly to obtain a globally more optimal packing. In addition, imposing such constraints explicitly leads to inferior performance. We found that the performance (average reward) drops about $20\%$ when adding such constraints, as shown in right of Figure~\ref{fig:scale}. 
This can also be verified by the experiment in Table~\ref{tab:boundary}.

\begin{table}
	\centering
	\caption{Evaluating the effect of boundary
	rule.}\label{tab:boundary}
	\scalebox{0.8}{
		\begin{tabular}{c|c|c}
			\whline{1.15pt}
			{} & {Space uti.} & {\# packed items} \\
			\whline{0.65pt}
			w/o corner \& boundary rule & $66.9\%$  & $17.5$            \\
			w corner \& boundary rule                 & $60.9\%$  & $16.2$        \\
			\whline{1.15pt}
	\end{tabular}}
\end{table}

\begin{figure}
	\centering
	\includegraphics[width=\linewidth]{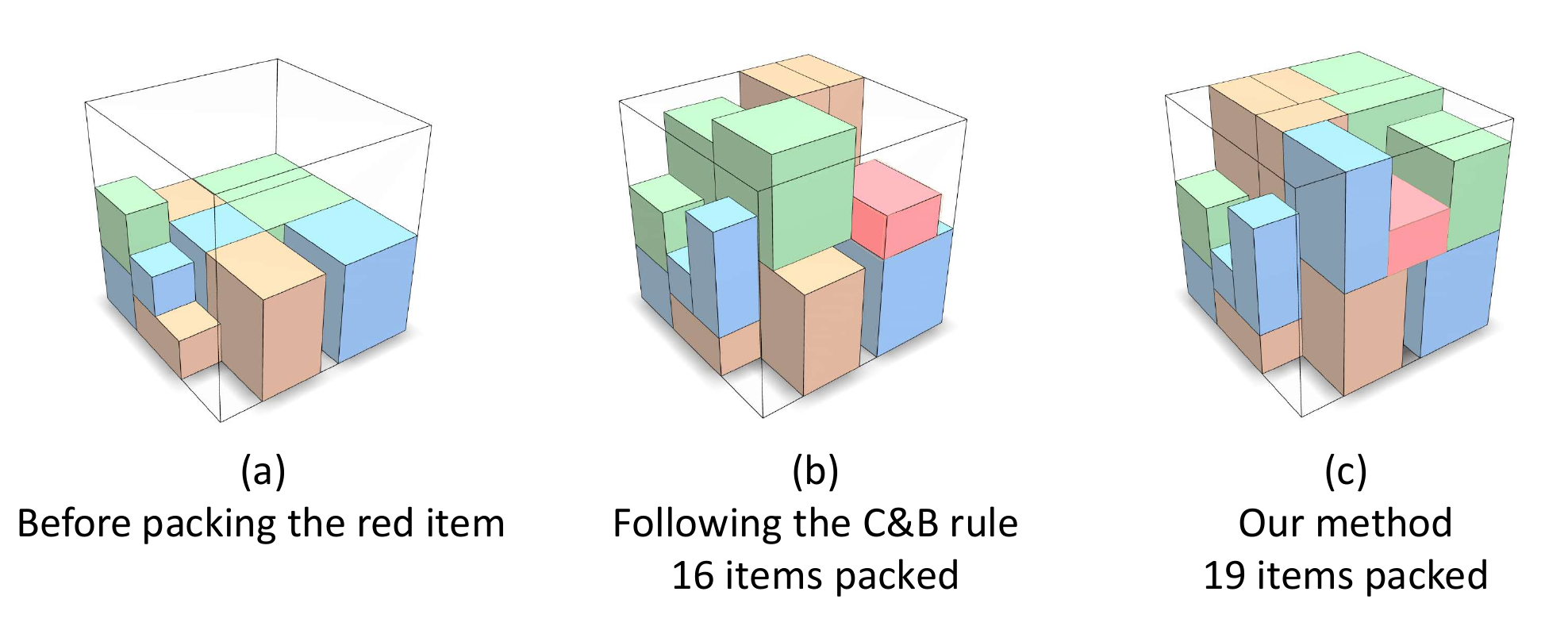}
	\caption{Visual comparison between learned vs. heuristic strategy. Different
packing strategy for the red item would lead to different performance.}
	\label{fig:boundary2}
\end{figure}

\begin{table}
	\centering
	\caption{Space utilization of unseen items.}\label{tab:unseen}
	\scalebox{1.2}{
		\begin{tabular}{c|c|c}
			\whline{1.15pt}
			{} & {$64 \rightarrow 64$}
& {$40 \rightarrow 64$} \\
			\whline{0.65pt}
			RS & $50.5\%$
& $49.4\%$  \\
			\whline{1.15pt}
	\end{tabular}}
\end{table}

To illustrate why our agent can decide when to follow the
C\&B rule to obtain a globally more optimal packing, we give a visual example here. As shown in Figure~\ref{fig:boundary2} (b), if the
agent exactly follow the C\&B rule when packing the red
item, it will leave gaps around the item. However, our method (Figure~\ref{fig:boundary2} (c)) can make a decision of packing the item in the middle upon the yellow and blue ones. Our method is trained to consider the whether there is enough room for next moves but not only takes the current situation into consideration. This move reserves enough space around the red item for the following item and this decision makes our method packing 3 more items when dealing with a same sequence.

\paragraph{Generalizability with unseen items}
We also test our method with unseen items. In this experiment, we randomly choose 40 items from the pre-defined item set $\mathcal{I}$ to train an agent and test it in complete $\mathcal{I}$. 
All the items are generated with RS and their test dimensions may not be seen during training.

The result is presented in Table~\ref{tab:unseen}. It shows that our method does demonstrate some generalizability and provides a reasonable benchmark. 



\begin{figure}[t]
	\centering
	\includegraphics[width=\linewidth]{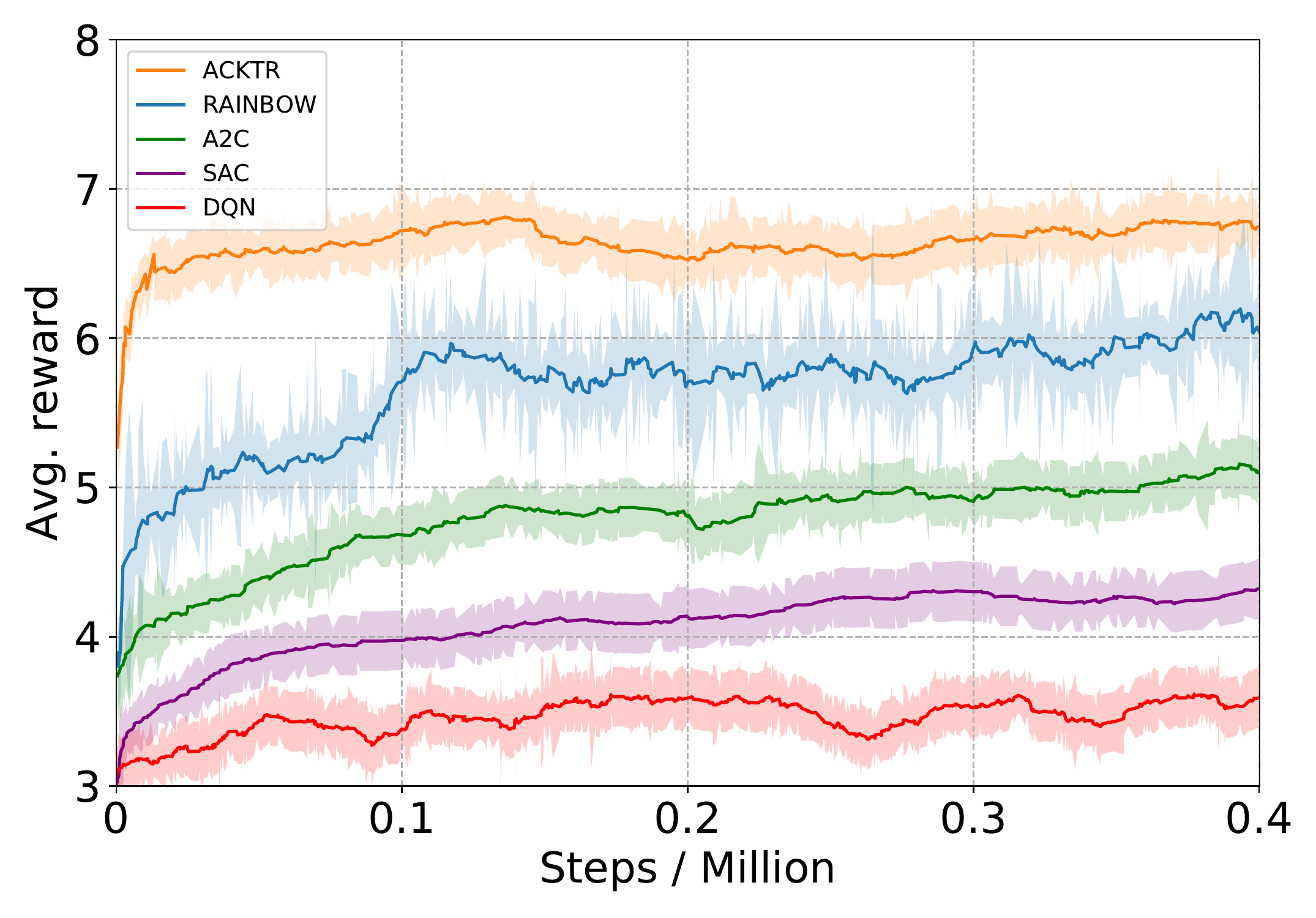}
	\caption{Performance of different DRL frameworks on CUT-2.}
	\label{fig:rl}
\end{figure}

\paragraph{Generalizability with untrained sequences}

Since our RS, CUT-1 and CUT-2 benchmarks are constructed based on different types of sequences, we can also evaluate the performance of our method on different sequence type from the training data. The result is presented in Table~\ref{tab:cro}, our method can still perform well while testing on varied sequences. Note that our model trained on CUT-2 attains the best generalization since this benchmark has the best balance between variation and completeness.


\paragraph{Different DRL framework}
We also test our method with different DRL frameworks on CUT-2 dataset with well-tuned parameters. For on-policy methods, we have evaluated A2C~\cite{mnih2016asynchronous} and ACKTR~\cite{wu2017scalable}. And we also evaluated DQN~\cite{mnih2015}, RAINBOW~\cite{HesselMHSODHPAS18} and SAC~\cite{haarnoja2018soft} for off-policy methods.
Figure~\ref{fig:rl} and Table~\ref{tab:drl} demonstrate that ACKTR can achieve the fastest convergence speed and best performance.

\begin{figure}[h]
	\centering
	\includegraphics[width=\linewidth]{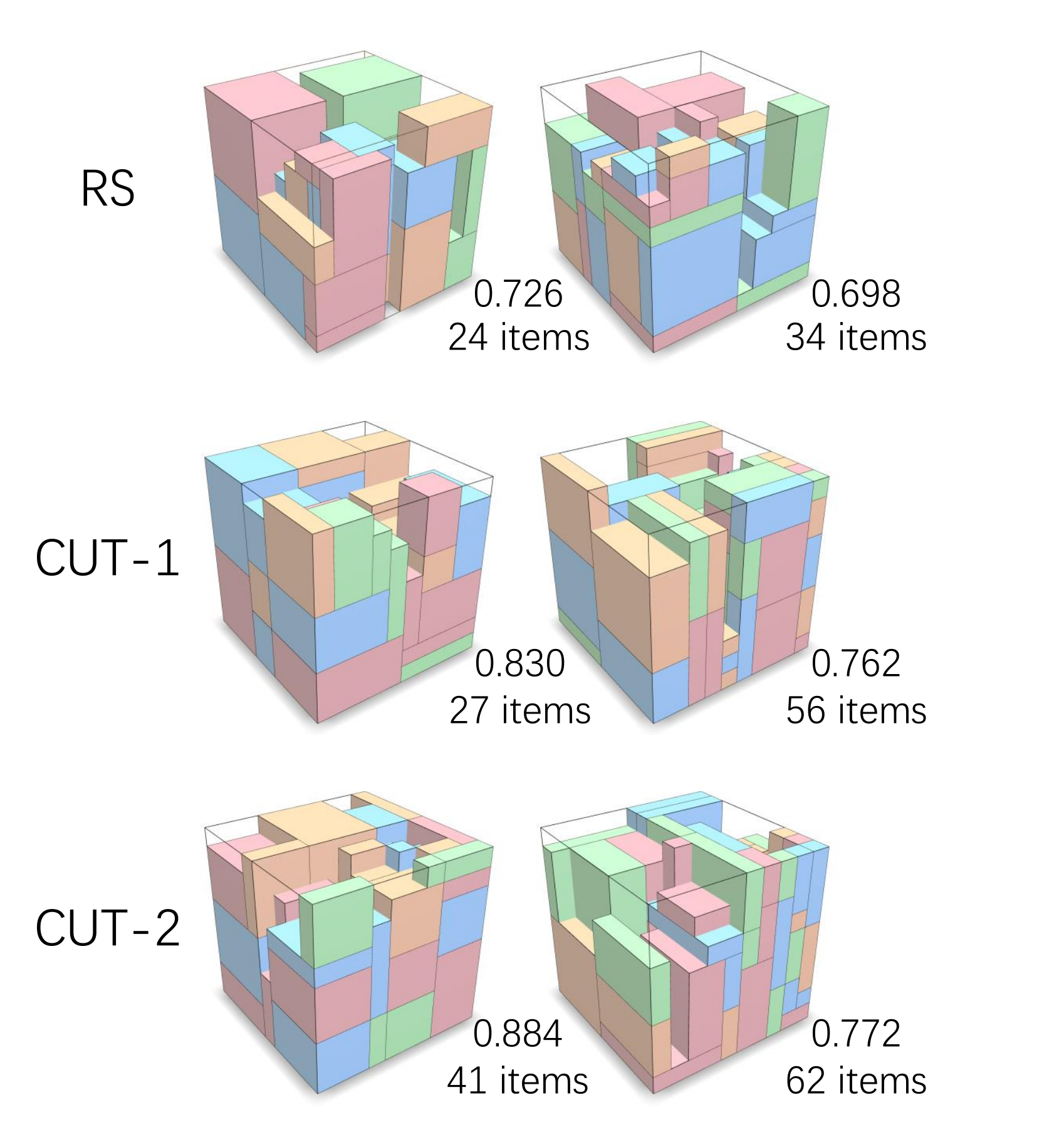}
	\caption{Packing results of our BPP-$1$ model with 125 types of pre-defined items.}
	\label{fig:125results}
\end{figure}

\begin{table}[h]
	\centering
	\caption{Evaluation of generalizability  with  untrained  sequences.}\label{tab:cro}
	\scalebox{1.0}{
		\begin{tabular}{c|c|c|c}
			\whline{1.15pt}
			{Test} & {Train} & {Space utilization} & {\#Packed items} \\
			\whline{0.65pt} 
			\multirow{3}*{RS} & RS & $50.5\%$  & $12.2$\\
			~&CUT-1& $43.4\%$&$10.7$\\
			~&CUT-2& $47.6\%$&$11.6$\\
			\whline{0.65pt}
			\multirow{3}*{CUT-1} & RS & $60.8\%$  & $15.7$\\
			~&CUT-1& $\bf{73.4\%}$&$\bf{19.1}$\\
			~&CUT-2& $69.4\%$&$17.9$\\
			\whline{0.65pt}
			\multirow{3}*{CUT-2} & RS & $60.9\%$  & $16.1$\\
			~&CUT-1& $62.4\%$&$16.6$ \\
			~&CUT-2& $\bf{66.9\%}$&$\bf{17.5}$ \\		
				\whline{1.15pt}
			\end{tabular}}\\
\end{table}

\paragraph{Study of increasing item dimensions}
We add more item dimensions to our pre-defined item set $\mathcal{I}$ and $|\mathcal{I}|$ is enlarged to 125. The newly added items also satisfy the condition that $l \leq L/2$, $w \leq W/2$, and $h \leq H/2$ to ensure the complexity of BPP
, which also means more little items has been added (one of item's axes must be 1). The result can be seen from Table~\ref{tab:little} and Figure~\ref{fig:125results}.

\clearpage
\begin{figure*}
{\centering
\begin{minipage}{0.95\columnwidth}
   \centering
	\captionof{table}{Performance on $|\mathcal{I}|=125$.}\label{tab:little}
	\scalebox{1.2}{
		\begin{tabular}{c|c|c}
		   \whline{1.15pt}
			{} & {Space uti.} & {\# items} \\ 
			\whline{0.65pt}
		   RS & $46.3\%$  	  & $18.3$   \\ 
		   CUT-1 & $59.2\%$  & $21.0$ \\ 
		   CUT-2 & $58.3\%$  & $22.1$ \\ 
			\whline{1.15pt}
		   \end{tabular}}\\
\end{minipage}\hspace{30 pt}
\begin{minipage}{0.95\columnwidth}
	\vspace{15 pt}
	\centering
	\captionof{table}{Performance of different DRL frameworks on CUT-2.}\label{tab:drl}
	\scalebox{1.2}{
		\begin{tabular}{c|c|c}
		   \whline{1.15pt}
			{DRL} & {Space uti.} & {\# items} \\ 
			\whline{0.65pt}
			ACKTR       & $66.9\%$  & $17.5$ \\
		   RAINBOW		& $58.8\%$  & $15.5$ \\
		   A2C  		& $53.0\%$  & $13.6$ \\
		   SAC  		& $44.2\%$  & $11.8$ \\
		   DQN  		& $35.3\%$  & $9.3$ \\
			\whline{1.15pt}
		   \end{tabular}}\\
\end{minipage}
}
\end{figure*}

\clearpage

\input{MCTS.tex}
\input{multibin.tex}

\input{benchmark.tex}

%% file: MCTS.tex
\SetKwInput{KwInput}{Inputs}                
\SetKwInput{KwOutput}{Output}              

\newcommand{\argmaxF}{\mathop{\mathrm{argmax}}\limits}
\begin{algorithm}
	\caption{Permutation MCTS}\label{alg:mcts}
    \SetKwProg{Fn}{Function}{:}{}
    \SetKwFunction{SEARCH}{SEARCH}
    \Fn{\SEARCH{$s_0$}}{
	    Create root node $v_0$ with state $s_0$
	    \While{$t<T$}
	    {
	        Copy $I$ as R\;
		    Sort R according to original order\;
		    $v_l,R \gets {\mbox{TREEPOLICY}}(v_0, R)$\;
		    $\Delta \gets {\mbox{DEFAULTPOLICY}}(v_l.s, R)$\;
		    $\mbox{BACKUP}(v_l,\Delta)$\;
		    $t \gets t+1$\;
	    }
		\While{$v.item$ is not $i_0$}
		{
		    $v \gets {\mbox{BESTCHILD}}(v,0)$\;
		}
		\Return $v.a$\;
    }
    \SetKwFunction{TREE}{TREEPOLICY}
    \Fn{\TREE{$v, R$}}{
		    \While{$v.s$ is non-terminal}
		    {
		        \If{v not fully expanded then}
		        {
		            \Return $\mbox{EXPAND}(v,R)$\;
		        }
		        \Else{
					$R \gets R\backslash v.item$\;
		            $v \gets \mbox{BESTCHILD}(v,c)$\;
		            
		        }
		    }
		\Return $v, R$\;
    }
    \SetKwFunction{DEFAULT}{DEFAULTPOLICY}
    \Fn{\DEFAULT{$s, R$}}
    {
        $eval = 0$\;
        \While{$s$ is non-terminal or $R$ is not empty}{
            $i \gets$ first item in R\;
		    $a \gets \pi(s,i)$\;
		    $s \gets SimEnv(s,i,a)$\;
		    \If {$s$ is non-terminal }
		    {
		        $eval \gets eval + Reward(i)$\;
		    }
		    \Else{
                $R \gets R\backslash i$\;
            }
            \If {$R$ is empty}
            {
                $eval \gets eval + V(s,Last)$\;
            }
            \Return eval\;
        }
    }
    \SetKwFunction{EXP}{EXPAND}
    \Fn{\EXP{$v,R$}}
    {
        Choose $i \in $ unused item from $R$\;
		Add new child $v'$ to $v$\\
		\quad with $v'.a = \pi(v.s,i)$\\
		\quad with $v'.s = SimEnv(v.s,i,v'.a)$\\
		\quad with $v'.item = i$\;
		$R \gets R\backslash i$\;
		\Return $v',R$\;
    }
    \SetKwFunction{BEST}{BESTCHILD} 
    \Fn{\BEST{$v,c$}}
    {
        \Return $\argmaxF_{v' \in children(v)}(v'.Q+c{\sqrt{\frac{v.N}{1+v'.N}}})$\;
    }
    \SetKwFunction{BACK}{BACKUP} 
    \Fn{\BACK{$v,\Delta$}}
    {
        \While {$v$ is not null}
        {
            $v.N \gets v.N+1$\;
		    $v.Q \gets max(v.Q,\Delta)$\;
		    $v \gets$ parent of $v$\;
        }
    }
\end{algorithm}

%% file: multibin.tex
\begin{algorithm}
  \caption{Bin Selecting Algorithm}\label{alg:multibin}
    \KwIn {The current item $n$, the set of candidate bins $B$\;
    }
    \KwOut {The most fitting bin $b_{best}$\;
    }
    Initialize the set of bin scores $B.val$ with $s_{def}$\;
    $b_{best} \gets \mathop{\arg\max}_{b \in B}V(H(b),n) - b.val$\;
    $b_{best}.val \gets value(H(b_{best}),n)$\;
    \Return $b_{best}$\;
\end{algorithm}

%% file: benchmark.tex
\SetKwInput{KwInput}{Inputs}                
\SetKwInput{KwOutput}{Output}              

\begin{algorithm}
	\caption{Benchmark Construction}\label{alg:construction}
	\label{alg:obb}
	\KwInput{valid item size threshold $(l_{min}, w_{min}, h_{min})$ $\sim  (l_{max}, w_{max}, h_{max})$, bin size $(L,W,H)$\;}
	
	\SetKwFunction{BEN}{Construction of pre-defined items collection}
    \SetKwProg{Fn}{Function}{:}{}
    \Fn{\BEN{$F$}}{
	    Initialize invalid item list $\mathcal{L}_{invalid} = \{ (L, W, H)\}$, valid item list $\mathcal{L}_{valid}= \emptyset$\;
		\While{$\mathcal{L}_{invalid} \neq \emptyset$}
		{
		    Randomly pop an $item_i$ from $\mathcal{L}_{invalid}$\;
		    Randomly select an axis $a_i$ of $item_i$,
		    which $a_i \textgreater a_{max} , a_i \in \{ x_i, y_i, z_i \}$\;
			Randomly split the $item_i$ into two sub items along axis $a_i$\;
			Calculate sub items' FLB corner coordinate $(lx, ly, lz)$\;

			\For{$item \in item_{sub}$}{
		    \If{$a_{min} \leq a_{sub} \leq a_{max}$}
		    {
		        Add the $item$ into $\mathcal{L}_{valid}$\;
		    }
		    \Else{
		        Add the $item$ into $\mathcal{L}_{invalid}$\;
			}
			}
		}
		\Return $\mathcal{L}_{valid}$\;
	}
    \SetKwFunction{CCC}{CUT-1}
    \Fn{\CCC{$\mathcal{L}_{valid}$}}{
		    Initialize items sequence $S=  \emptyset $\;
		    Sort $\mathcal{L}_{valid}$ by $lz_i$ coordinate of each item in ascending order\;
		    $s_i \gets item_i$'s index in the sorted list\;
		    \Return $S$\;
    }
	\SetKwFunction{CC}{CUT-2}
    \Fn{\CC{$\mathcal{L}_{valid}$}}{
		Initialize height map $\mathcal{H}_n \in Z^{L \times W}$\; 
		$\mathcal{H}_n = 0^{L \times W}, S = \emptyset$\;
	    \While{$\mathcal{L}_{valid} \neq \emptyset$}
	    {
	        Randomly pop an $item_i$ from $\mathcal{L}_{valid}$ satisfy $$lz_i = \mathcal{H}_n(item_i)$$
			
			Add the $item_i$ into $S$\;
		    $\mathcal{H}_n(item_i) \gets\mathcal{H}_n(item_i) + h_i $\;
	    }
		\Return $S$\;
    }
\end{algorithm}